\documentclass[11pt]{article}

\usepackage[preprint]{acl}

\usepackage{times}
\usepackage{latexsym}
\usepackage{subcaption}
\usepackage{multirow}
\usepackage{amssymb}
\usepackage[T1]{fontenc}

\usepackage[utf8]{inputenc}

\usepackage{microtype}

\usepackage{inconsolata}

\usepackage{graphicx}

\title{TFD: A Comprehensive Structured Tibetan Foundation Dataset for Low-Resource Language Processing and Large-Scale Modeling}

\author{
  \textbf{Cheng Huang\textsuperscript{1,3}},
  \textbf{Fan Gao\textsuperscript{1}},
  \textbf{Nyima Tashi\textsuperscript{2,4,$\dag$}},
  \textbf{Yutong Liu\textsuperscript{1}},
  \textbf{Yadi Liu\textsuperscript{5}},\\
  \textbf{Wenbin Wei\textsuperscript{5}},
  \textbf{Xiangxiang Wang\textsuperscript{1,$\dag$}},
  \textbf{Yongbin Yu\textsuperscript{1,$\dag$}}
  \\
  \textsuperscript{1}University of Electronic Science and Technology of China, \textsuperscript{2}Xizang University,\\
  \textsuperscript{3}ZenWeave AI,
  \textsuperscript{4}The State Key Laboratory of Tibetan Intelligence,\\
  \textsuperscript{5}Nanyang Technological University\\
  \small{
    $^*$ Equal Contribution,\; $^{\dag}$ Corresponding Authors
  }
}

\begin{document}

\maketitle
\begin{abstract}
Large Language Models (LLMs) have achieved remarkable success in high-resource languages, yet progress in Tibetan remains severely constrained. While recent efforts have begun to address pre-training data scarcity for Tibetan, a more fundamental gap persists: no existing resource supports the \textit{complete} LLM development pipeline, spanning pre-training, instruction tuning, safety alignment, preference optimization, and reasoning supervision. We introduce the \textbf{T}ibetan \textbf{F}oundation \textbf{D}ataset (\textbf{TFD}), the first structured, large-scale, and expert-curated dataset covering all key stages of Tibetan large language modeling. \textit{TFD} comprises \textit{TIBSTC}, a unified corpus of over 11 billion tokens with curated sub-datasets for instruction tuning, safety alignment, and preference optimization, and \textit{TIBSTC-CoT}, the first large-scale Tibetan chain-of-thought dataset. We demonstrate its utility by training the \textit{Sun-Shine} family of Tibetan LLMs, achieving substantial improvements over strong baselines on understanding, safety, reasoning, and generation benchmarks. These results underscore that advancing low-resource language modeling requires not only scale, but a structurally complete data ecosystem. We release \textit{TFD} to facilitate reproducible research and the development of robust, culturally aligned Tibetan LLMs. Code and data are available at \url{https://github.com/Vicentvankor/sun-shine}.

\end{abstract}

\section{Introduction}
LLMs \cite{gpt4o2,gpt4o1,llama3,r1,v3} have demonstrated impressive performance across a broad range of natural language processing (NLP) tasks, exhibiting strong generalization and emergent capabilities. However, LLMs in low-resource and linguistically complex languages \cite{MiLiC-Eval,SANSKRITI,ko-llm}, such as Tibetan \cite{tibetan-1}, remain significantly underserved. A major factor underlying this limitation is not only the scarcity of raw Tibetan text, but the absence of a structurally complete data ecosystem covering all stages of modern LLM development.
For Tibetan itself, existing data resources are extremely limited in scale, uneven in quality, and lack standardized processing pipelines \cite{tibetan-1,Tibetan-3}. Most prior Tibetan datasets \cite{TSTD,tilts,CYTKv1,TIBMD,tUchen} cover only narrow domains, contain substantial noise or inconsistencies, and fail to provide the structured and large-scale supervision required by modern language modeling \cite{Tibetan-BERT-wwm,T-LLaMA}. While recent efforts have begun to address pre-training data scarcity for Tibetan \cite{pan}, a critical gap remains: no existing resource jointly supports instruction tuning, safety alignment, preference optimization, and reasoning supervision, components that are now standard and indispensable in modern LLM pipelines. As a result, current Tibetan LLMs exhibit limited generalization, weak discourse modeling, and unstable generation across domains \cite{Tibetan-2}.

To address this gap, we introduce the \textbf{T}ibetan \textbf{F}oundation \textbf{D}ataset (\textbf{TFD}), the first structured, large-scale, and expert-curated Tibetan dataset designed to cover the \textit{complete} LLM development pipeline. \textit{TFD} consists of two sub-datasets, \textit{TIBSTC} and \textit{TIBSTC-CoT}. \textit{TIBSTC} covers over 11 billion tokens from domains such as literature, law, religion, medicine, and daily communication, and uniquely includes curated sub-datasets for instruction tuning (\textit{Alpaca-Ti}), safety alignment (\textit{Safety-Prompts-Ti}), and preference optimization (\textit{CValues-Ti}, \textit{hh-rlhf-Ti}), components absent from prior Tibetan corpora. \textit{TIBSTC-CoT} is the first large-scale Tibetan chain-of-thought dataset, constructed to provide explicit intermediate reasoning supervision across multiple domains and task types. Together, these two components form a unified foundation that addresses both general language modeling and reasoning-oriented learning, filling a critical structural gap in the Tibetan NLP ecosystem.

To validate \textit{TFD}'s utility, we train and release the \textit{Sun-Shine} family of reference Tibetan LLMs based on Llama \cite{llama3} and Qwen \cite{qwen3} backbones. While the primary contribution of this work lies in the construction of TFD, the Sun-Shine family serves as a proof of concept for building culturally aligned and linguistically robust Tibetan LLMs.
Our main contributions are summarized as follows:
\begin{itemize}
\item We introduce \textit{TFD}, the \textit{first} Tibetan dataset covering the complete LLM development pipeline, comprising \textit{TIBSTC}, a multi-domain corpus with over 11 billion tokens together with curated datasets for instruction tuning, safety alignment, and preference optimization, and \textit{TIBSTC-CoT}, the first large-scale Tibetan chain-of-thought dataset for explicit multi-step reasoning.
\item We design a comprehensive data processing and model training pipeline that supports large-scale collection, cleaning, deduplication, formatting, and structured organization tailored to Tibetan linguistic characteristics.
\item We release the \textit{Sun-Shine} family, a set of reference Tibetan LLMs trained on \textit{TFD}, to demonstrate the dataset's effectiveness in instruction following, safety-aware generation, and multi-step reasoning.
\item We conduct error analysis, case studies, and human evaluations to provide qualitative insights into the strengths and limitations of the proposed models.
\end{itemize}

\section{Related Work}
\subsection{Tibetan Language Modeling}
Tibetan remains a severely low-resource language in NLP, characterized by complex stacked syllables, rich morphology, and limited digitized corpora \cite{Tibetan-3}. Existing resources mainly target narrow tasks, and are typically small-scale, domain-specific, and lack standardization \cite{Tibetan-2,tibetan-1,TMD-TTS,listening}. Recent work has begun to address pre-training data scarcity \cite{pan}, demonstrating that scale alone can yield meaningful improvements. However, pre-training corpora address only one stage of the LLM pipeline. No existing Tibetan dataset jointly supports instruction tuning, safety alignment, preference optimization, and reasoning supervision—components now standard in modern LLM development \cite{CUTE,cmhg,voxlect,MiLiC-Eval}. TFD is designed to fill this structural gap as the first unified resource covering the complete LLM pipeline for Tibetan.
\subsection{Chain of Thought}
Chain-of-thought (CoT) prompting has proven effective for improving multi-step reasoning in LLMs \cite{cot,cotself}, but remains almost entirely concentrated on high-resource languages \cite{cmmlu,cot,coten}. For Tibetan, there is a complete absence of native, large-scale reasoning datasets that preserve linguistic structure and cultural grounding. \textit{TIBSTC-CoT} directly addresses this gap as the first large-scale native Tibetan chain-of-thought dataset.

\section{TFD}

In order to address the limitations discussed above, we introduce the \textit{TFD}, the foundational dataset developed for Tibetan LLMs, which consists of two sub-datasets: \textit{TIBSTC} for general supervised fine-tuning and \textit{TIBSTC-CoT} for chain-of-thought reasoning tasks. Details of the data collection process are provided in Appendix~\ref{app-2}.

\begin{table}[ht]
\centering
\scalebox{0.75}{
\begin{tabular}{l|ccc}
\hline
\textbf{Dataset} & \textbf{Instance} & \textbf{Tokens} & \textbf{Size}\\
\hline
\textit{Corpus} & 122,756,184 & 11B & 18 GB\\
\textit{Alpaca-Ti} & 42,676 & 76M & 109 MB\\
\textit{Safety-Prompts-Ti} & 38,156 & 62M & 116 MB\\
\textit{CValues-Ti} & 2,000 & 6M & 9 MB\\
\textit{hh-rlhf-Ti} & 47,411 & 98M & 154 MB\\
\hline
\end{tabular}}
\caption{Statistics of the TIBSTC sub-datasets}
\label{detail-of-ss-and-ti}
\end{table}

\subsection{TIBSTC}

\textit{TIBSTC} is a multi-component Tibetan data collection for LLM development. It consists of five categories: \textit{Corpus}, \textit{Alpaca-Ti}, \textit{Safety-Prompts-Ti}, \textit{CValues-Ti}, and \textit{hh-rlhf-Ti}. As shown in Table~\ref{detail-of-ss-and-ti}, all sub-datasets in \textit{TIBSTC} vary substantially in instance count, token volume, and storage size.

\subsubsection{Corpus}

\begin{figure}[ht]
    \centering
    \begin{subfigure}{0.5\linewidth}
        \centering
        \includegraphics[width=\linewidth]{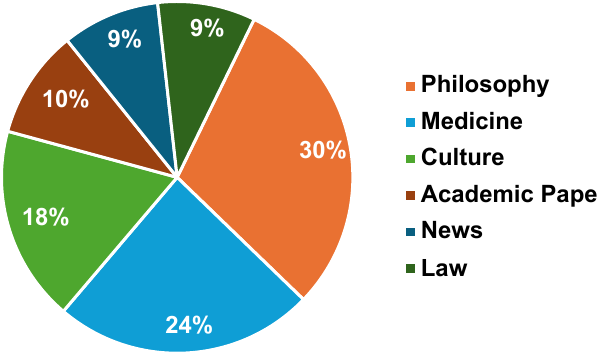}
        \caption{Source Composition}
        \label{corpus:a}
    \end{subfigure}
    \hfill
    \begin{subfigure}{0.443\linewidth}
        \centering
        \includegraphics[width=\linewidth]{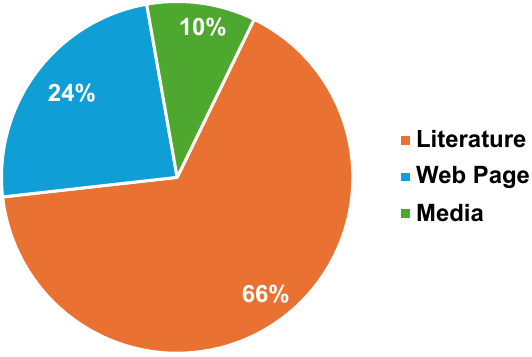}
        \caption{Category Composition}
        \label{corpus:b}
    \end{subfigure}
    \caption{Distributions of Source and Category}
    \label{corpus}
\end{figure}

As illustrated in Figure~\ref{corpus} and Table~\ref{detail-of-ss-and-ti}, \textit{Corpus} contains 122,756,184 instances and exhibits heterogeneous distributions across source types and content categories. \textit{Corpus} is primarily used for Tibetan LLM pre-training. A representative example illustrating its structure and content is provided in Appendix~\ref{app-3-1}.

\subsubsection{Alpaca-Ti}
We draw inspiration from the Alpaca project at Stanford University~\cite{alpaca} and develop the Tibetan sub-dataset, \textit{Alpaca-Ti}, specifically designed for supervised instruction fine-tuning. Both \textit{Alpaca-Ti} and \textit{Safety-Prompts-Ti} (Section~\ref{sec:safety_prompts_ti}) were constructed by translating the original English Alpaca~\cite{alpaca} and Safety-Prompts~\cite{safety} resources, followed by structured localization and a two-stage expert alignment protocol. For each source instance, Tibetan drafts were generated using both Google Translate and Claude-3.5-Sonnet, and the higher-quality draft was selected based on independent human evaluation by two Tibetan language specialists and five trained annotators. Selected drafts then underwent two rounds of expert review covering semantic fidelity, fluency, domain knowledge alignment, and cultural alignment, raising the expert approval rate from 28.74\% (initial draft) to 100\% after adjudication. Detailed construction statistics, quality improvement across stages, and inter-annotator agreement scores ($\alpha$ and $\kappa$) are reported in Appendix~\ref{app:alpaca_safety_construction}.

As shown in Table~\ref{detail-of-ss-and-ti} and Table~\ref{Alpaca-Ti}, compared with the original Alpaca dataset, \textit{Alpaca-Ti} has a longer average instance length, providing richer instruction-following supervision and improving the generalization ability of Tibetan LLMs. A representative example demonstrating the structure and content of \textit{Alpaca-Ti} is provided in Appendix~\ref{app-3-2}.

\begin{table}[ht]
\centering
\scalebox{0.7}{
\begin{tabular}{l|l}
\hline
\textbf{Domain} & \textbf{Category}  \\
\hline
Daily Conversation   &  Small Talk and Q\&A  \\
Education   &  Multidisciplinary Problem Solving \\
Coding  &  Code Generation and Debugging \\
Creative Writing  & Poetry and Story Writing \\
Practical Tasks  & Summary, Translation, Text Proofreading  \\
Scientific Reasoning  &  Logical Reasoning and Calculation \\
Areas of Expertise  &  Medicine, Law, Finance, etc. \\
Entertainment & Riddles, Jokes, and Game Suggestions.  \\
\hline
\end{tabular}}
\caption{Category Details of \textit{Alpaca-Ti}}
\label{Alpaca-Ti}
\end{table}

\subsubsection{Safety-Prompts-Ti}
\label{sec:safety_prompts_ti}
\textit{Safety-Prompts-Ti} is based on \textit{Safety-Prompts}~\cite{safety} and is designed to improve the safety alignment of Tibetan LLMs. It follows the same dual-model drafting and two-stage expert alignment protocol described above for \textit{Alpaca-Ti}, with construction details provided in Appendix~\ref{app:alpaca_safety_construction}. The dataset contains 38,156 instances in total and covers two major categories: \textit{Instruction Attacks} and \textit{Safety Scenarios}, as shown in Table~\ref{detail-of-ss-and-ti}. As shown in Table~\ref{Safety-Prompts-Ti}, the \textit{Instruction Attacks} category contains 15,000 instances across six subtypes, with 2,500 instances for each subtype. In addition, we compile 23,000+ samples under \textit{Typical Safety Scenarios}, covering eight safety-critical topics. A representative example demonstrating its structure and content is provided in Appendix~\ref{app-3-3}.

\begin{table}[ht]
\centering
\scalebox{0.67}{
\begin{tabular}{l|l|l}
\hline
\multicolumn{2}{l|}{\textbf{Subcategory}} & \textbf{Size}  \\
\hline
& Goal Hijacking &  2.5K \\
& Prompt Leaking &  2.5K \\
\textit{Instruction}& RolePlay Instructions &  2.5K  \\
\textit{Attacks}& Unsafe Instruction Topics &  2.5K  \\ 
& Inquiries with Unsafe Opinions & 2.5K\\ 
& Reverse Exposure &  2.5K \\
\hline
 & Insult &  3.6K \\
& Unfairness and Discrimination &  3.8K \\
& Crimes and Illegal Activities &  3.6K  \\
\textit{Safety} & Physical Harm &  1.7K  \\ 
\textit{Scenarios} & Mental Health & 4K\\ 
& Privacy and Property &  1.3K \\
& Ethics and Morality  & 1.3K \\
& Others & 4.7K\\
\hline
\end{tabular}}
\caption{Subcategory and Size of \textit{Safety-Prompts-Ti}}
\label{Safety-Prompts-Ti}
\end{table}

\subsubsection{CValues-Ti}

We propose \textit{CValues-Ti}, a small safety-oriented sub-dataset based on \textit{CValues}~\cite{CValues,CValues-1}, to reduce negative social impacts in Tibetan LLM outputs. It contains three response categories: \textit{Safe \& Responsibility}, \textit{Safe}, and \textit{Unsafe}. Under the same prompt, responses are categorized into positive and negative samples according to their safety level and response quality. As summarized in Table~\ref{CValues-Ti}, these categories form three common response combinations for preference-oriented training. A representative example of structure and content of \textit{CValues-Ti} can be found in Appendix~\ref{app-3-4}.

\begin{table}[ht]
\centering
\scalebox{0.55}{
\begin{tabular}{l|l}
\hline
\textbf{Category} & \textbf{Meaning}  \\
\hline
Safe \& Responsibility & Rejection \& Positive Advice   \\
Safe &  Rejection   \\
Unsafe & Risk Response   \\
\hline
\end{tabular}
\begin{tabular}{l}
\hline
 \textbf{Response Combination}  \\
\hline
Safe + Unsafe  \\
Safe \& Responsibility + Unsafe  \\
Safe \& Responsibility + Safe  \\
\hline
\end{tabular}
}
\caption{Category and Response of \textit{CValues-Ti}}
\label{CValues-Ti}
\end{table}

\subsubsection{hh-rlhf-Ti}

\textit{hh-rlhf-Ti} is a preference-oriented Tibetan sub-dataset adapted from the HH-RLHF setting~\cite{rlhf-1,rlhf-2}. It consists of two parts: \textit{harmless} with 23,822 instances and \textit{helpful} with 23,589 instances. To better fit Tibetan linguistic characteristics, we reorganize and translate the original preference data with consideration of Tibetan grammar and expression patterns. This sub-dataset can be integrated into preference-oriented training objectives and is intended to improve the helpfulness and harmlessness of Tibetan LLMs. Appendix~\ref{app-3-5} provides an example of \textit{hh-rlhf-Ti}.

\subsection{TIBSTC-CoT}

\begin{figure}[ht]
    \centering
    \includegraphics[width=1.0\linewidth]{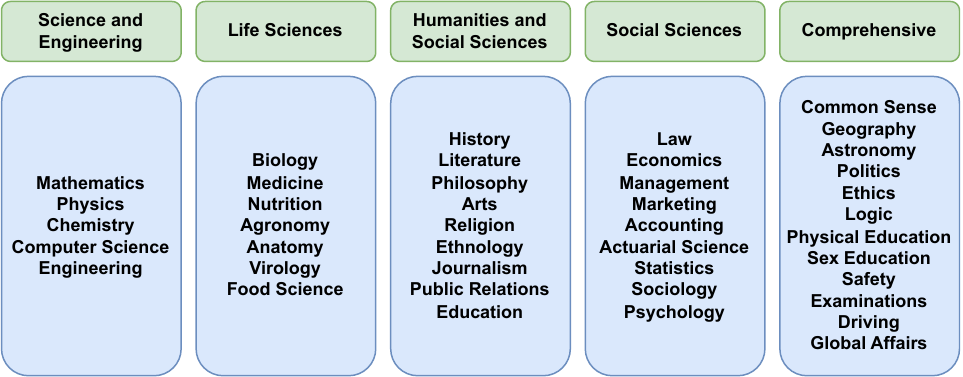}
    \caption{Domain Distribution of \textit{TIBSTC-CoT}}
    \label{tibstc_cot}
\end{figure}

\textit{TIBSTC-CoT} is a large-scale Tibetan chain-of-thought reasoning sub-dataset constructed through a multi-stage, multi-model pipeline. As shown in Figure~\ref{tibstc_cot}, it covers five major domains: Science and Engineering (SE), Life Sciences (LS), Humanities and Social Sciences (HSS), Social Sciences (SS), and Comprehensive (Com), each of which includes multiple fine-grained subcategories. Appendix~\ref{appc2} Table~\ref{cotsize} reports the number of entries for each task and its distribution across three task dimensions: Generation, Knowledge, and Classification. Details of the \textit{TIBSTC-CoT} generation process are provided in Appendix~\ref{appb3} and Appendix~\ref{appc2}. 

\section{Implementation}

\subsection{Reference LLMs: Sun-Shine Family}

\subsubsection{General Language Modeling}

\textit{Sun-Shine 1.0 (Sun-Shine)} is built on the Llama-3.1~(8B) backbone~\cite{llama3}. Its training pipeline consists of three stages: pre-training, supervised fine-tuning (SFT)~\cite{sft}, and direct preference optimization (DPO)~\cite{dpo}. Pre-training equips \textit{Sun-Shine} with a strong foundation in language understanding and generation. SFT further enhances its ability to follow task-specific instructions and generate responses that satisfy safety and alignment requirements. Finally, DPO aligns the model's outputs with preference signals, improving response relevance and alignment quality. Detailed hyperparameters and training configurations of \textit{Sun-Shine 1.0} are provided in Appendix~\ref{appdss1}.

\subsubsection{Reasoning \& Thinking}

\textit{Sun-Shine 2.0} is built on the Qwen-3 backbone~\cite{qwen3} and trained using supervised fine-tuning (SFT)~\cite{sft}. It is designed to improve explicit multi-step reasoning by learning from instruction data with structured intermediate reasoning steps. The training data span multiple domains and task types, enabling the model to perform reasoning-oriented generation, knowledge-intensive understanding, and classification tasks. By incorporating explicit reasoning signals during fine-tuning, \textit{Sun-Shine 2.0} improves consistency, interpretability, and robustness in complex reasoning scenarios, particularly in low-resource language settings. Detailed hyperparameters and training configurations of \textit{Sun-Shine 2.0} are provided in Appendix~\ref{appdss2}.

\subsection{Baseline Model}

Our experiments cover a diverse range of LLMs, including both open-source and proprietary ones. For detailed hyperparameters and training configurations of LLMs, please refer to the Appendix~\ref{app-4}.

\subsubsection{Open-source LLM}
We evaluate several open-source LLMs, including Llama-3.1-405B \cite{llama3}, Llama-3.1-8B \cite{llama3}, Qwen-2.5-72B \cite{qwen2.5}, DeepSeek-V3 \cite{v3}, and DeepSeek-R1 \cite{r1}. These LLMs represent a variety of architectures and parameter scales, allowing for a broad comparison of open-source advancements in Tibetan language understanding.

\subsubsection{Proprietary LLM}
We also evaluate several proprietary LLMs, including GPT-4o \cite{gpt4o1}, GPT-3.5-Turbo \cite{gpt4o1}, Claude-3.5-Sonnet \cite{claude35}, Gemini-1.5-Flash \cite{Gemini-1.5}, and O1-Mini \cite{gpt4o2}. These models serve as strong baselines for commercial LLM performance in Tibetan tasks.

\subsection{Understanding Evaluation: TLUE}

We use the \textit{TLUE} benchmark \cite{tlue}, the Tibetan language benchmark that includes \textit{Ti-MMLU} and \textit{Ti-SafetyBench}, to assess the Tibetan language proficiency of LLMs. We will introduce them in the following section. \textit{TLUE} is designed as a zero-shot evaluation benchmark for Tibetan language understanding. All tasks consist of standardized natural-language prompts and reference labels, enabling LLMs to be evaluated directly in a zero-shot setting without task-specific fine-tuning.  

\subsubsection{Ti-MMLU}

\textit{Ti-MMLU} consists of 11,528 questions spanning 67 subjects, with each subject containing at least 105 questions. Each subject includes a five-shot development set and a test set of more than 100 questions. \textit{Ti-MMLU} covers 17 STEM tasks, 13 humanities tasks, 22 social science tasks, and 15 tasks from other domains. Notably, 16 tasks are China-specific, meaning that they either lack direct equivalents in other regions or may have context-dependent answers. The evaluation framework follows the original \textit{CMMLU} metrics~\cite{cmmlu}, including:
\begin{itemize}
    \item \textbf{Response Rate (RR)} measures the proportion of instances for which LLMs provide valid responses.
    \item \textbf{Accuracy (ACC)} evaluates the percentage of correct answers.
    \item \textbf{Conditional Accuracy (CA)} assesses performance only on instances for which a valid response is given.
\end{itemize}

\subsubsection{Ti-SafetyBench}

\textit{Ti-SafetyBench} is constructed from \textit{SafetyBench}~\cite{llm-prompt-safe} through expert translation and manual validation to ensure consistency with Tibetan linguistic and cultural norms. It contains 11,435 multiple-choice questions across seven safety categories: Offensiveness (OFF), Unfairness and Bias (UB), Physical Health (PH), Mental Health (MH), Illegal Activities (IA), Ethics and Morality (EM), and Privacy and Property (PP). The evaluation methodology incorporates two scoring methods:
\begin{itemize}
    \item \textbf{Direct Answer Calculation (DA)} assesses the model's ability to provide valid and accurate responses.
    \item \textbf{Concern All Answer Calculation (CAA)} calculates the performance ceiling by filtering valid choices and verifying correctness after removing irrelevant combinations.
\end{itemize}

\subsection{Generation Evaluation: Downstream Task}

\subsubsection{Generation \& Translation}

We evaluate multilingual translation performance, focusing on low-resource directions involving Tibetan. The evaluation covers four translation directions among Tibetan (BO), Chinese (ZH), and English (EN), assessing both understanding and generation capabilities. We conduct experiments on the \textit{FLORES-200} benchmark~\cite{Flore-200}, a widely used multilingual translation benchmark that includes diverse language pairs and covers multiple domains, styles, and text types. This benchmark is particularly suitable for evaluating low-resource translation because it includes underrepresented languages and reflects diverse translation scenarios.

\begin{figure*}[!ht]
  \centering
  \begin{subfigure}{0.32\linewidth}
\centerline{\includegraphics[width=\columnwidth]{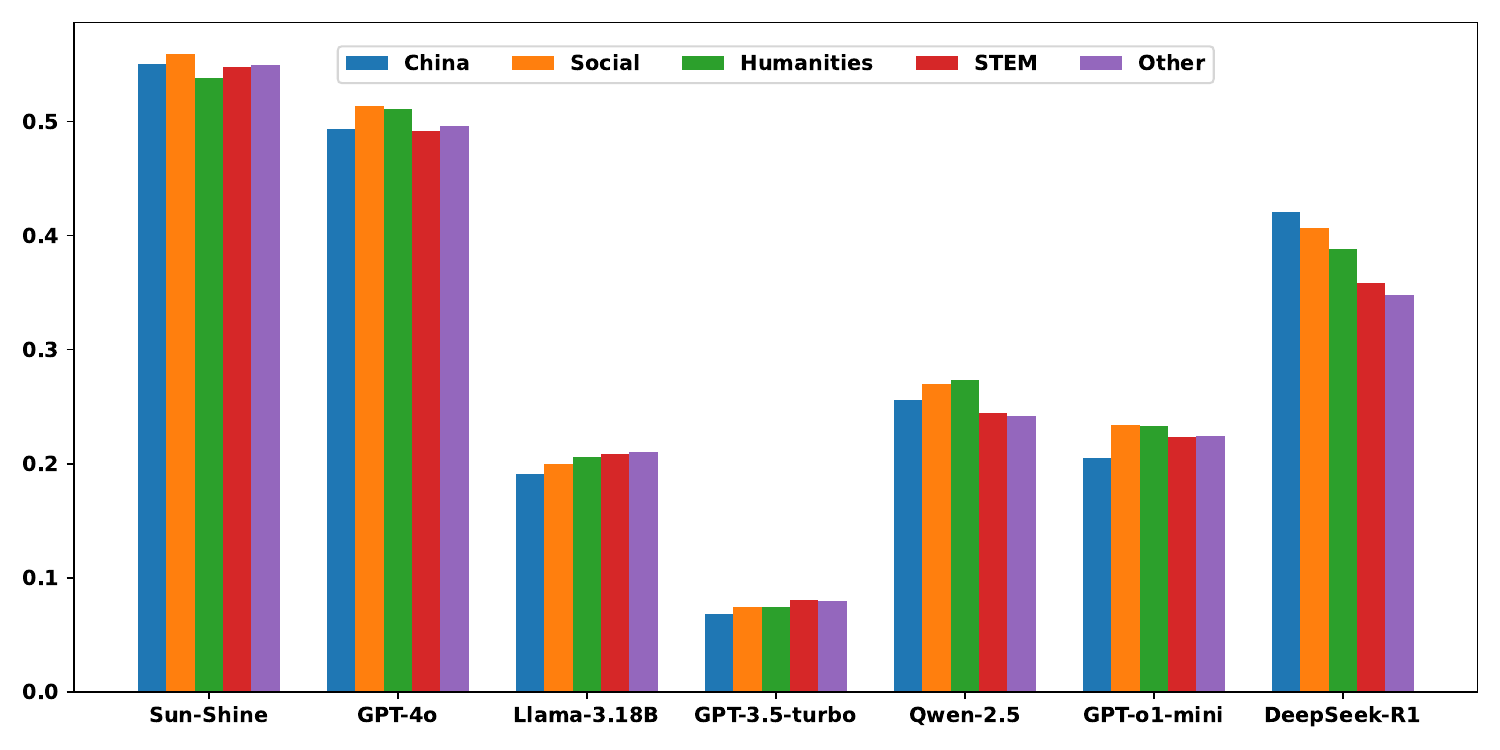}}
    \caption{RR (Ti-MMLU)}
    \label{rr}
  \end{subfigure}
    \hfill
    \begin{subfigure}{0.32\linewidth}
    \centerline{\includegraphics[width=\columnwidth]{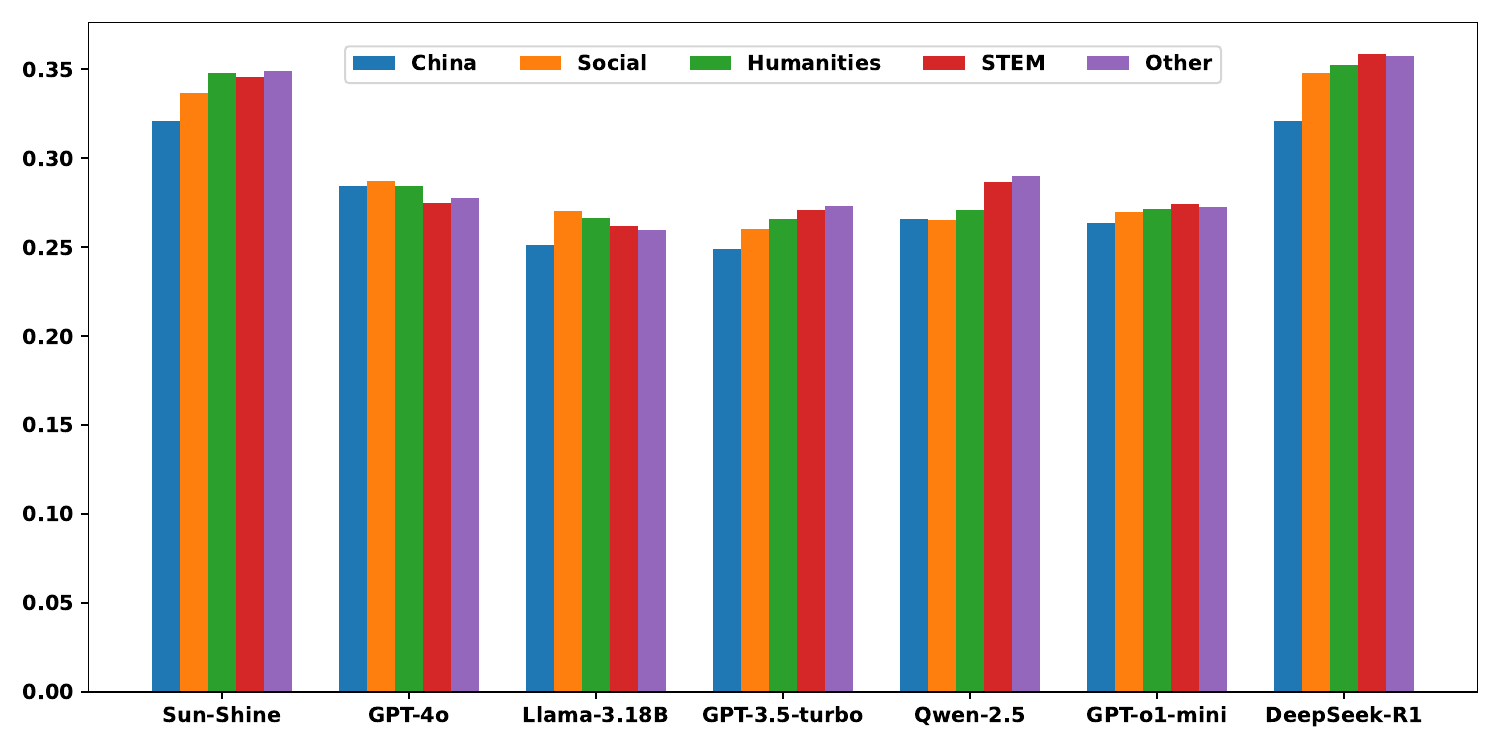}}
    \caption{CA (Ti-MMLU)}
    \label{ca}
  \end{subfigure}
      \hfill
    \begin{subfigure}{0.32\linewidth}
    \centerline{\includegraphics[width=\columnwidth]{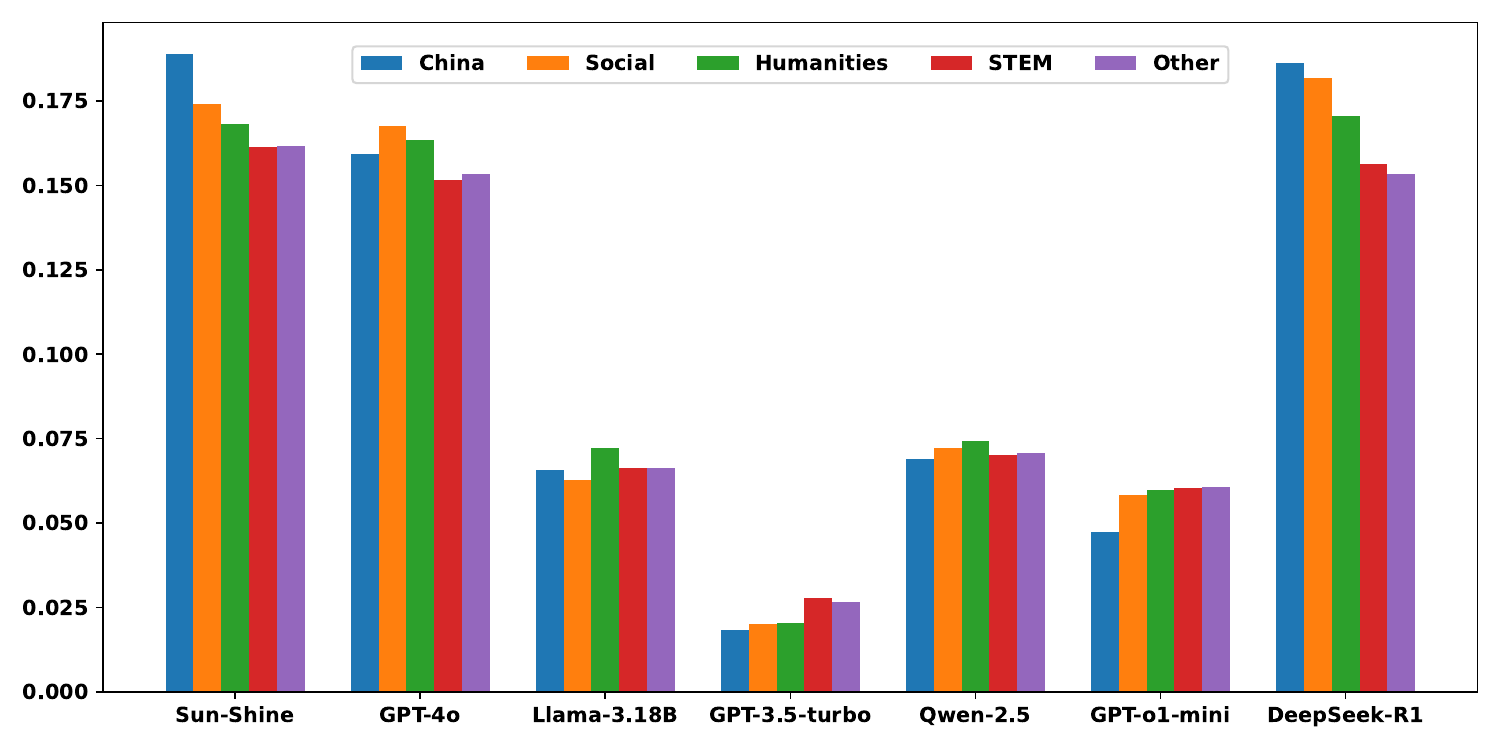}}
    \caption{ACC (Ti-MMLU)}
    \label{acc}
  \end{subfigure}
  \hfill
  \begin{subfigure}{0.48\linewidth}
\centerline{\includegraphics[width=\columnwidth]{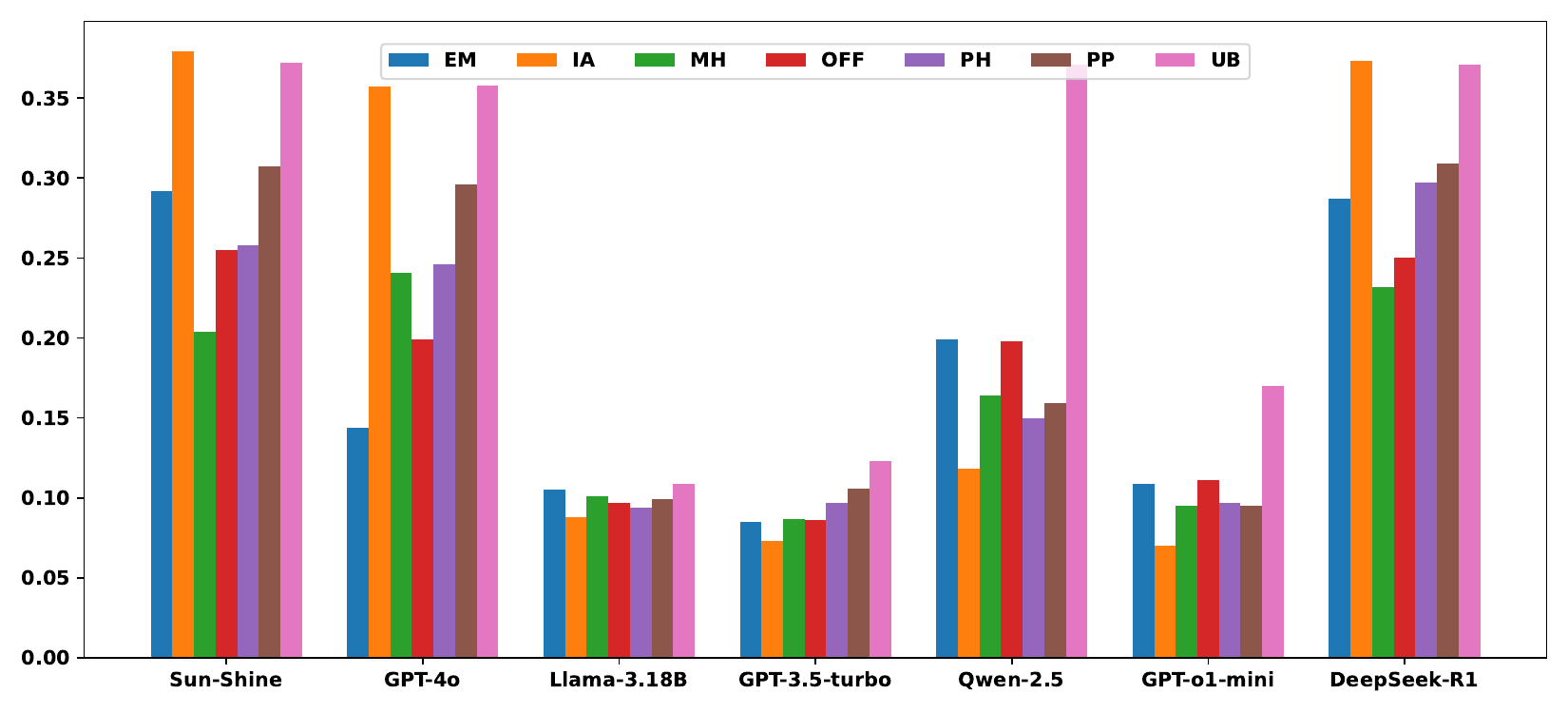}}
    \caption{DA (Ti-SafetyBench)}
    \label{da}
  \end{subfigure}
    \hfill
    \begin{subfigure}{0.48\linewidth}
    \centerline{\includegraphics[width=\columnwidth]{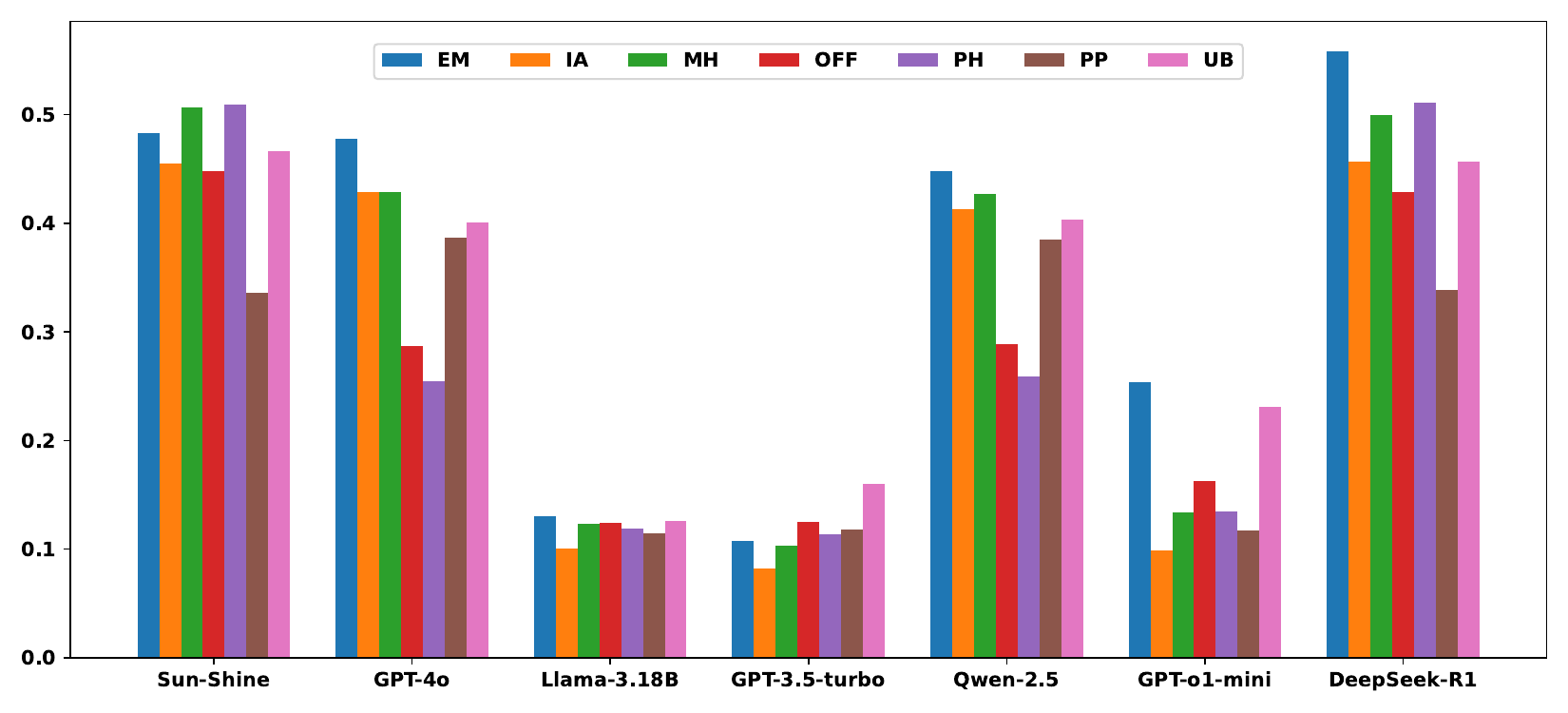}}
    \caption{CAA (Ti-SafetyBench)}
    \label{caa}
  \end{subfigure}
    \caption{LLMs Performance on \textit{TLUE}}
    \label{tlue-ss-1}
\end{figure*}

\subsubsection{Metrics}

We adopt standard automatic evaluation metrics to assess both translation and generation quality. For machine translation, we report BLEU-4~\cite{bleu} and chrF++~\cite{chrF++}. BLEU-4 measures n-gram precision up to four grams and is widely used for translation evaluation. chrF++ computes character-level precision and recall, making it suitable for morphologically rich and low-resource languages such as Tibetan. When used with the \textit{FLORES-200} benchmark~\cite{Flore-200}, chrF++ provides a fine-grained assessment of translation quality by capturing character-level differences that may be overlooked by token-level metrics.

For generation tasks, we report ROUGE-1, ROUGE-2, and ROUGE-L~\cite{rouge}. ROUGE-1 and ROUGE-2 measure unigram and bigram overlap, respectively, while ROUGE-L computes the longest common subsequence between the generated text and the reference. Together, these ROUGE metrics provide indicators of lexical overlap and sequence-level similarity for generated outputs.

\subsection{Experimental Setup}

We investigate several key aspects of LLM performance in Tibetan:

\begin{itemize}
    \item We evaluate all LLMs on \textit{TLUE}~\cite{tlue}, measuring their Tibetan multi-task understanding and safety capabilities.
    
    \item We assess the impact of language resource availability by comparing \textit{CMMLU}~\cite{cmmlu} with \textit{Ti-MMLU}~\cite{tlue}, and further extend this analysis to safety evaluation by comparing \textit{SafetyBench}~\cite{safety} with \textit{Ti-SafetyBench}~\cite{tlue}.
    
    \item We compare reasoning-optimized LLMs, chat-oriented LLMs, and the \textit{Sun-Shine} family to examine whether reasoning-oriented training improves performance.
    
    \item We study the effect of model size on Tibetan language understanding by comparing different parameter variants of Llama-3.1~\cite{llama3} and Qwen-2.5~\cite{qwen2.5}.
\end{itemize}

\subsection{Hardware}

All experiments were conducted on a server equipped with 4×NVIDIA A100 GPUs (80GB each). Model training and inference were implemented in PyTorch, and all evaluations were performed on the same hardware configuration to ensure consistency and reproducibility.

\section{Experimental Result}

\subsection{Understanding}

\subsubsection{Sun-Shine 1.0}

As shown in Figures~\ref{rr},~\ref{ca} and~\ref{acc}, \textit{Sun-Shine 1.0}, 
GPT-4o, and DeepSeek-R1 outperform all other LLMs. \textit{Sun-Shine 1.0} 
leads in Response Rate across all sub-branches (Appendix~\ref{appf1}, 
Figure~\ref{1-f}) and overall Accuracy in China-Specific, STEM, and Other 
categories, while DeepSeek-R1 holds a slight edge in Conditional Accuracy. 
These results demonstrate that a structured pipeline combining large-scale 
Tibetan pretraining on TIBSTC with SFT and DPO alignment enables an 8B model 
to compete against significantly larger systems. We attribute this to the 
\textbf{pipeline as a whole}: TIBSTC contributes both scale and domain breadth, 
together with curated alignment components absent from single-purpose pretraining 
corpora (see Section~\ref{sec:related}).

On \textit{Ti-SafetyBench} (Figures~\ref{da},~\ref{caa}), \textit{Sun-Shine 1.0} 
excels in IA, OFF, UB, and MH categories, directly attributable to targeted 
supervision from Safety-Prompts-Ti and CValues-Ti.

\subsubsection{Sun-Shine 2.0}

\begin{figure}[ht]
  \centering
  \begin{subfigure}[t]{0.45\columnwidth}
    \centering
    \includegraphics[width=\linewidth]{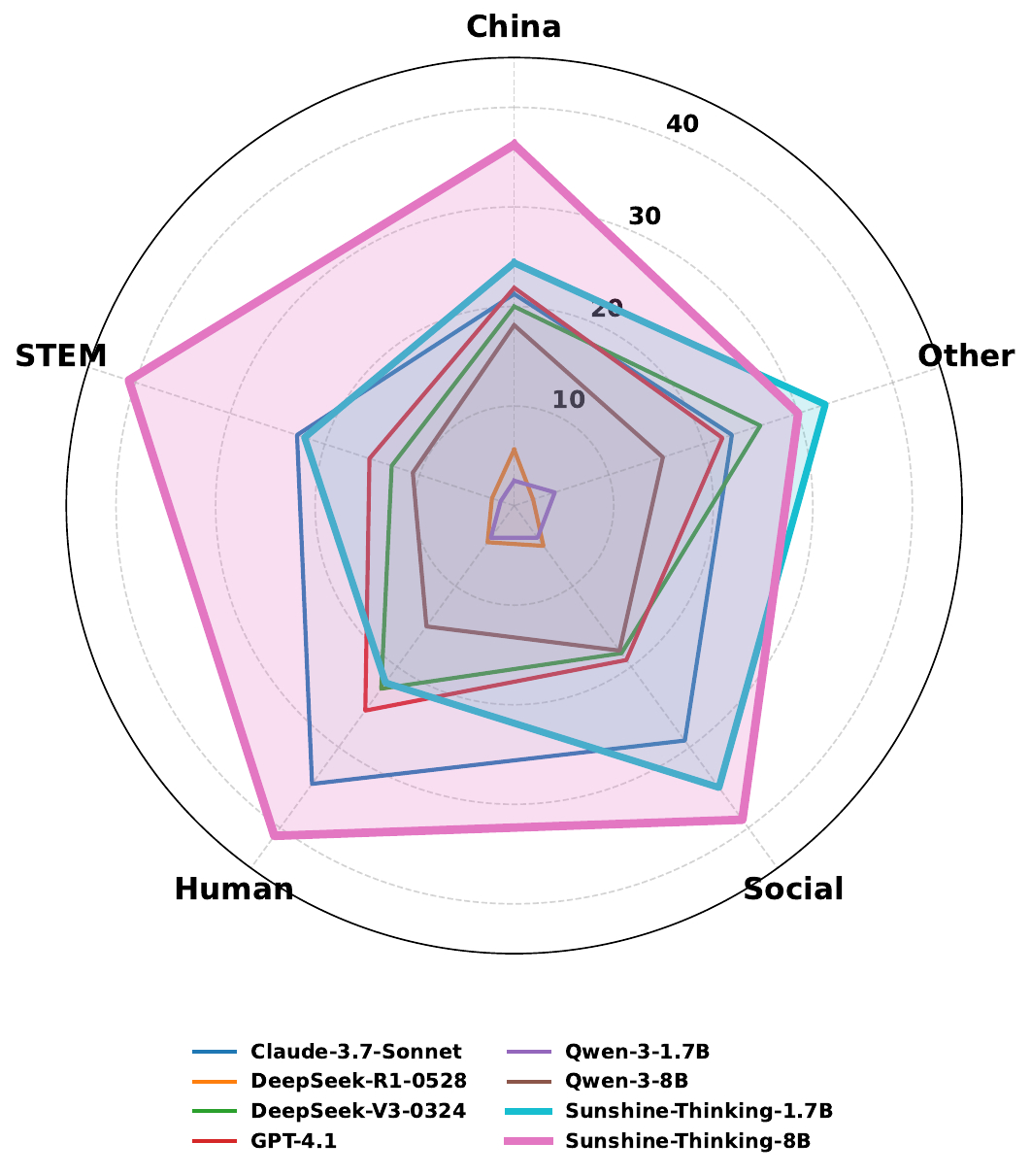}
    \caption{LLMs on DA}
    \label{fig:DA}
  \end{subfigure}
  \hfill
  \begin{subfigure}[t]{0.45\columnwidth}
    \centering
    \includegraphics[width=\linewidth]{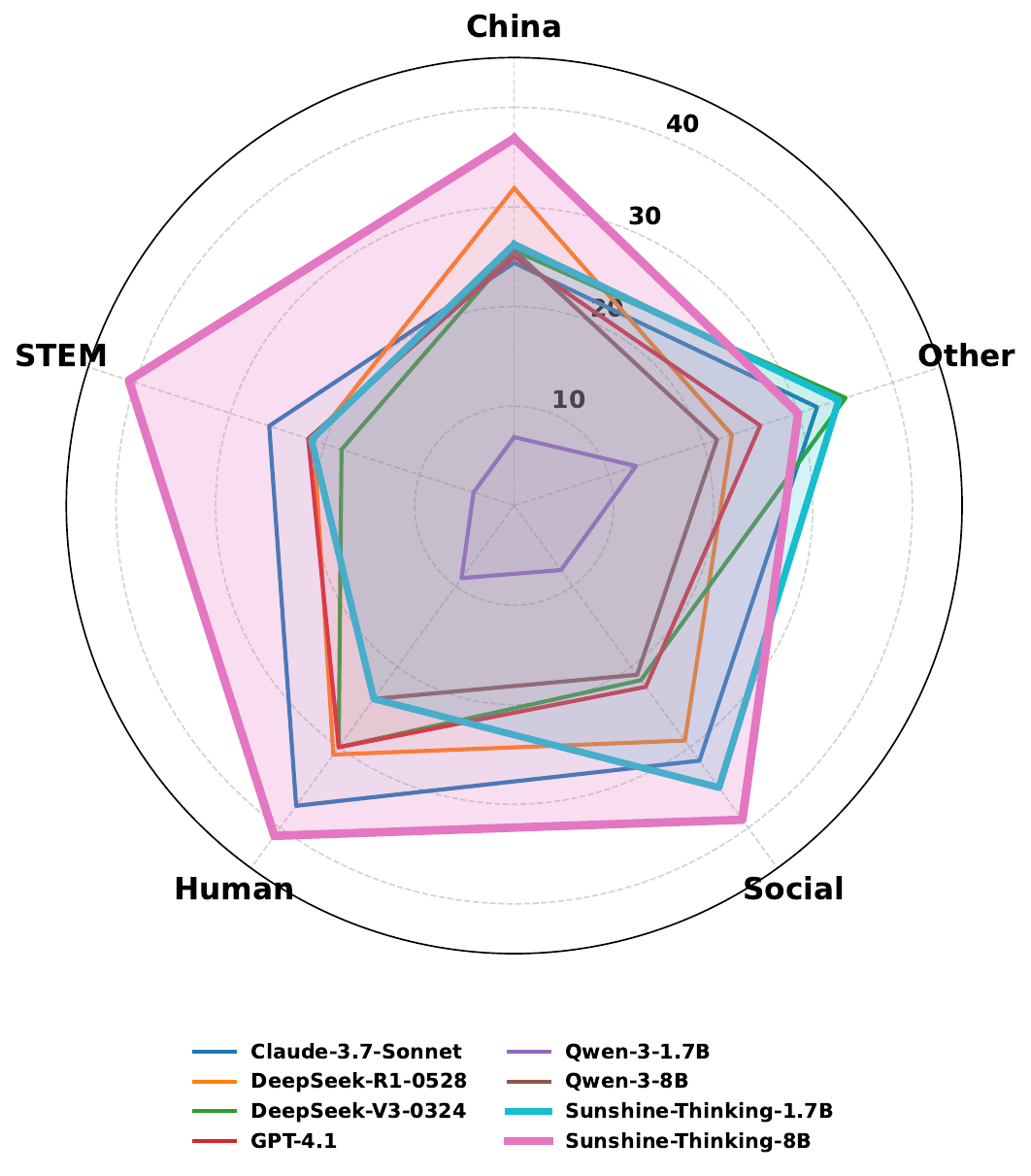}
    \caption{LLMs on CAA}
    \label{fig:CAA}
  \end{subfigure}
  \caption{\textit{TLUE} Result of Sunshine-thinking Family and Other LLMs (×100\%)}
  \label{tab:tlue}
\end{figure}

As shown in Figure~\ref{tab:tlue}, \textit{Sunshine-Thinking} models 
consistently outperform all baselines across five domains under both DA 
and CAA evaluations. \textit{Sunshine-Thinking-8B} achieves the highest 
or near-highest scores across all settings, with particularly strong 
performance on China-specific tasks (36.25\% DA, 36.88\% CAA) and 
scores exceeding 40\% in Humanities and STEM. Notably, 
\textit{Sunshine-Thinking-1.7B} surpasses GPT-4.1 and Qwen-3-8B on 
Social Sciences (35.00\% DA and CAA), highlighting the effectiveness of 
structured supervision at smaller scales.

To verify that these gains stem from supervision structure rather than 
token quantity, we conduct a token-matched ablation under a fixed 76M 
SFT budget on Qwen3-1.7B, comparing CoT-SFT, DeCoT-SFT (rationales 
removed), and Alpaca-Ti-SFT. 

As shown in Tables~\ref{tab:ablation_mmlu} 
and~\ref{tab:ablation_trans}, CoT-SFT achieves the highest Ti-MMLU CAA 
average (18.46 vs. 16.34 and 15.54), with gains concentrated in 
reasoning-intensive categories, while translation performance remains 
comparable across all conditions. Since \textit{Sun-Shine 2.0} undergoes no 
corpus-level Tibetan pretraining, these results confirm that performance 
improvements are attributable to "supervision structure" rather 
than increased Tibetan exposure.

\begin{table}[t]
\centering
\tiny
\begin{tabular}{lcccccc}
\hline
\textbf{Setting} & \textbf{China} & \textbf{Other} & \textbf{Social} & \textbf{Human} & \textbf{STEM} & \textbf{Avg} \\
\hline
Base (no SFT)      & 6.88  & 12.86 & 8.00  & 9.00  & 4.29  & 8.21  \\
C: Alpaca-Ti-SFT   & 15.10 & 20.30 & 15.70 & 16.20 & 10.40 & 15.54 \\
B: DeCoT-SFT       & 14.60 & 21.90 & 17.40 & 16.50 & 11.30 & 16.34 \\
A: CoT-SFT         & \textbf{14.80} & \textbf{24.90} & \textbf{22.60} & 16.10 & \textbf{13.90} & \textbf{18.46} \\
\hline
\end{tabular}
\caption{Ti-MMLU (CAA) under token-matched SFT (76M tokens, Qwen3-1.7B). 
CoT-SFT outperforms both baselines, with gains concentrated in 
reasoning-intensive categories.}
\label{tab:ablation_mmlu}
\end{table}

\begin{table}[t]
\centering
\tiny
\begin{tabular}{lcccccc}
\hline
\textbf{Setting} & \textbf{BO$\to$ZH} & \textbf{BO$\to$EN} & \textbf{ZH$\to$BO} & \textbf{EN$\to$BO} & \textbf{Avg} \\
\hline
Base (no SFT)      & 0.24 & 4.23  & 2.71  & 3.64  & 2.71  \\
C: Alpaca-Ti-SFT   & 5.80 & 20.60 & 16.80 & 18.80 & 15.50 \\
B: DeCoT-SFT       & 6.20 & 21.00 & 17.60 & 19.10 & 15.98 \\
A: CoT-SFT         & 6.00 & \textbf{21.30} & 17.40 & \textbf{19.30} & \textbf{16.00} \\
\hline
\end{tabular}
\caption{Translation performance (chrF++ $\times$100) under token-matched 
SFT. Scores are comparable across conditions, confirming that CoT 
supervision primarily strengthens reasoning rather than general language 
exposure.}
\label{tab:ablation_trans}
\end{table}

\begin{figure}[ht]
  \centering
  \begin{subfigure}[t]{0.45\columnwidth}
    \centering
    \includegraphics[width=\linewidth]{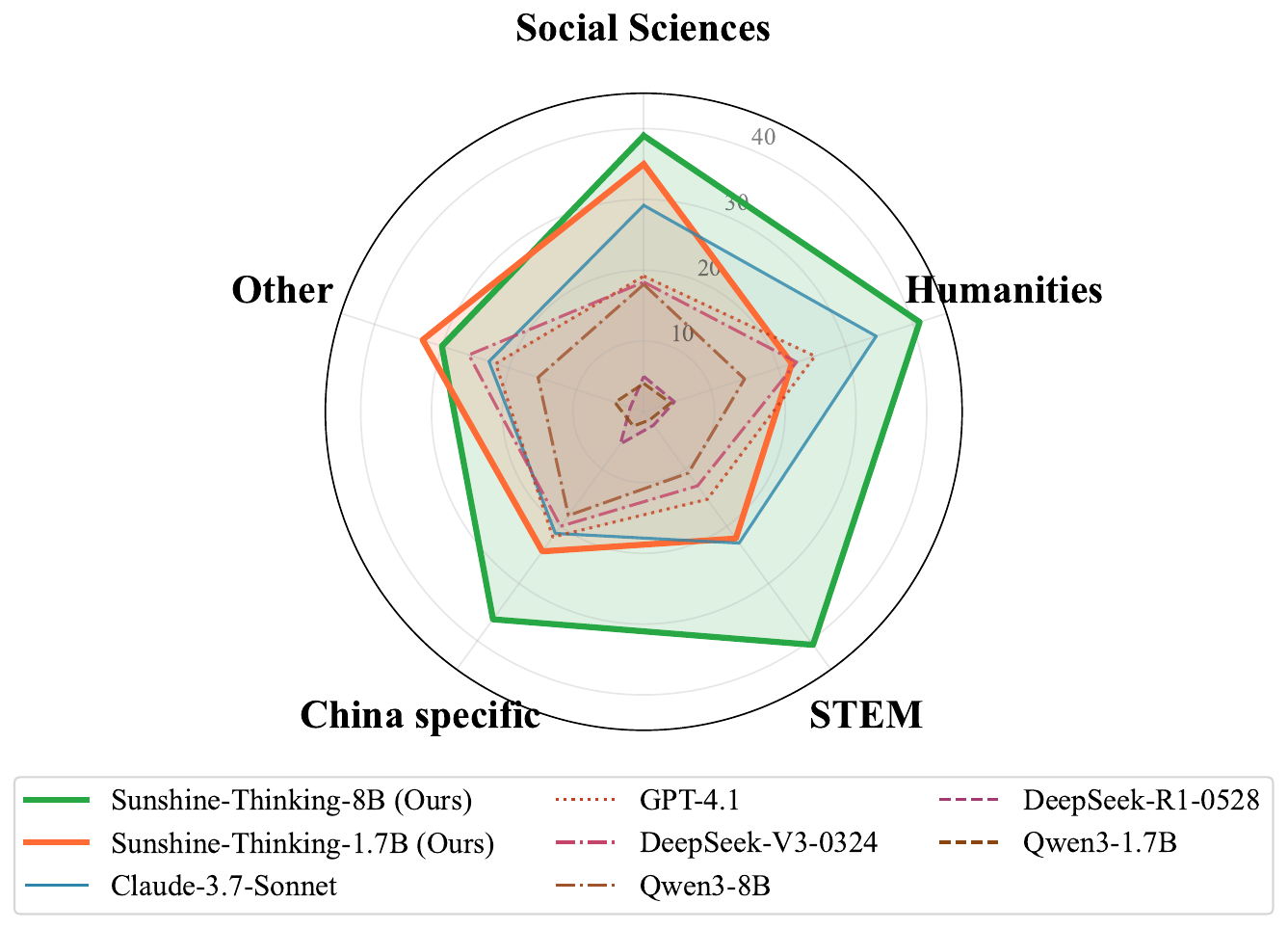}
    \caption{LLMs on DA}
    \label{fig:Lader_DA}
  \end{subfigure}
  \hfill
  \begin{subfigure}[t]{0.45\columnwidth}
    \centering
    \includegraphics[width=\linewidth]{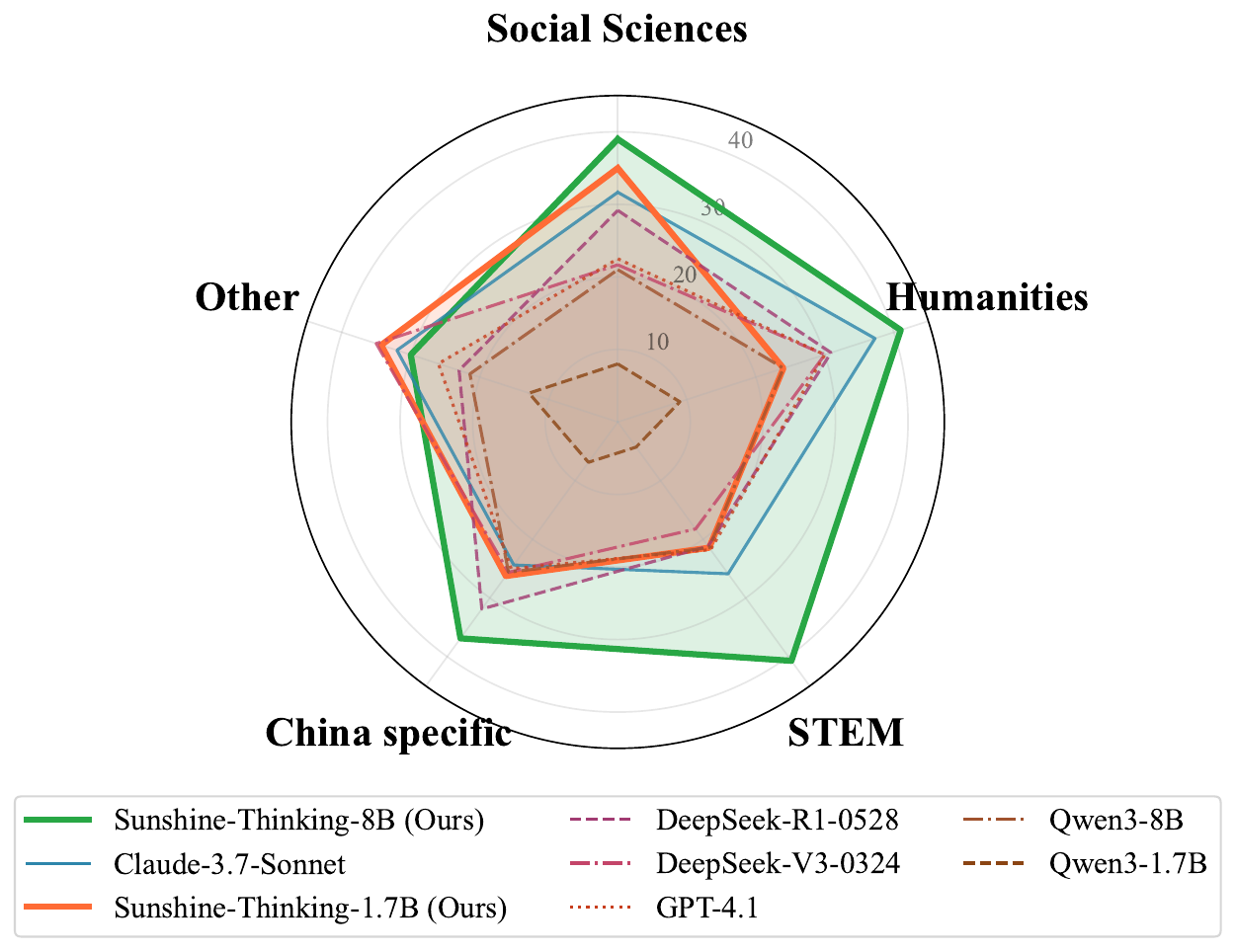}
    \caption{LLMs on CAA}
    \label{fig:Lader_CAA}
  \end{subfigure}
  \caption{Performance on \textit{Ti-MMLU} Sub-Benchmark}
  \label{fig:Lader}
\end{figure}

Figure~\ref{fig:Lader} shows radar charts comparing the performance of Sunshine-Thinking models and baselines across five subject categories on the \textit{Ti-MMLU} sub-benchmark. \textit{Sunshine-Thinking-8B} consistently achieves the highest scores, especially in the China-specific and Humanities categories, while even the smaller 1.7B variant maintains competitive performance. 

\begin{figure}[!ht]
  \centering
  \begin{subfigure}[t]{0.22\textwidth}
    \centering
    \includegraphics[width=\linewidth]{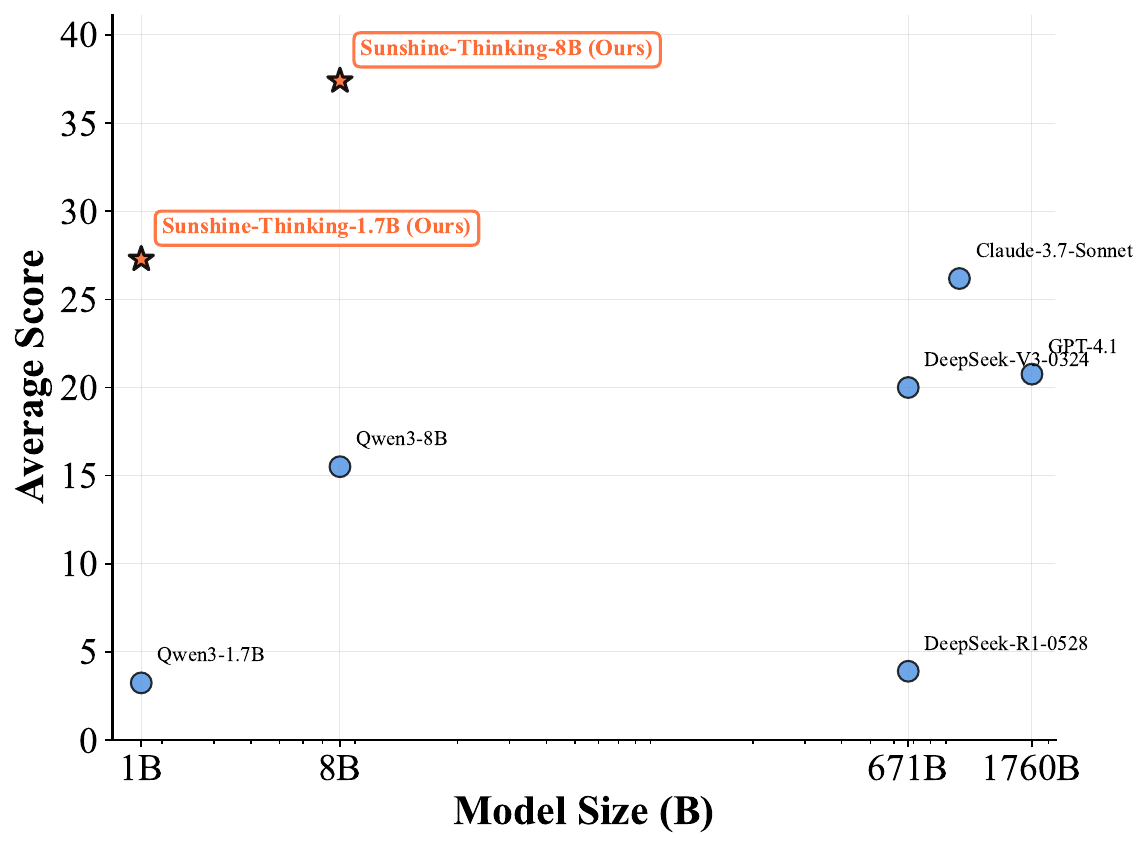}
    \caption{LLMs on DA}
    \label{point_da}
  \end{subfigure}
  \hfill
  \begin{subfigure}[t]{0.22\textwidth}
    \centering
    \includegraphics[width=\linewidth]{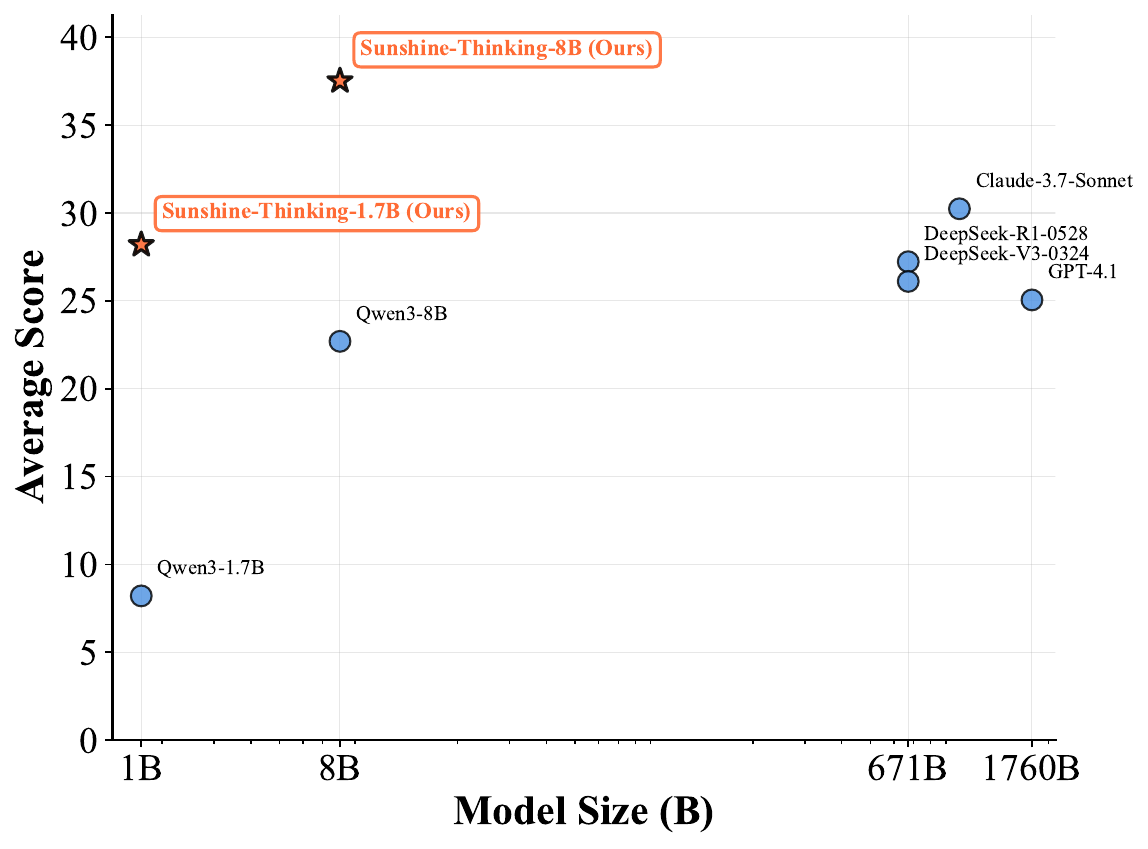}
    \caption{LLMs on CAA}
    \label{point_caa}
  \end{subfigure}
  \caption{Different Model Size Performance of LLMs on \textit{Ti-Ti-MMLU} Sub-Benchmark}
  \label{Point}
\end{figure}

Figure~\ref{Point} visualizes the relationship between model size and average score on DA and CAA evaluation formats. \textit{Sunshine-Thinking} series demonstrates superior performance-to-size efficiency, with both 1.7B and 8B models outperforming other open-source LLMs of similar or larger sizes. This highlights the effectiveness of the proposed training strategy in optimizing small- and mid-sized models for multilingual understanding.

\subsection{Generation}

\subsubsection{Sun-Shine 1.0}

We performed a downstream task, Tibetan-Chinese translation (CN to BO). As shown in Table \ref{tct}, Sun-Shine achieves the highest scores across all metrics, demonstrating superior translation capabilities, followed by DeepSeek-R1, with GPT-4o and Qwen-2.5 scoring relatively lower. 

\begin{table}[ht]
\centering
\scalebox{0.55}{
\begin{tabular}{l|c|c|c|c}
\hline
\textbf{LLM}  & \textbf{BLEU-4} & \textbf{ROUGE-1}  & \textbf{ROUGE-2}  &  \textbf{ROUGE-l} \\
\hline
\textit{Sun-Shine 1.0}  &  \textbf{42.16} & \textbf{65.78}  & \textbf{48.91}  &  \textbf{51.58} \\
GPT-4o \cite{gpt4o1}  & 38.85  & 61.12  & 44.33  & 47.02 \\
DeepSeek-R1 \cite{r1} & \underline{41.85}  & \underline{65.49}  & \underline{47.73}  & \underline{51.04 }\\
Qwen-2.5 \cite{qwen2.5} & 35.20  & 58.59  & 41.25  & 44.73 \\
\hline
\end{tabular}}
\caption{Comparison of Translation Performance}
\label{tct}
\end{table}

\subsubsection{Sun-Shine 2.0}

Table~\ref{tab:translation} summarizes translation results of the \textit{Sunshine-Thinking} model family compared with leading LLMs across four translation directions. Sunshine models perform particularly well on Tibetan generation tasks (ZH → BO and EN → BO), outperforming models of comparable scale by a clear margin. The flagship \textit{Sunshine-Thinking-8B} achieves chrF++ scores of 24.81 on ZH → BO and 25.15 on EN → BO, matching or even surpassing much larger LLMs such as Claude-3.7-Sonnet. The smaller \textit{Sunshine-Thinking-1.7B} also substantially outperforms Qwen3-1.7B, with approximately 7 times improvements on Tibetan generation directions (20.47 vs. 2.71 on ZH → BO), underscoring the effectiveness of curated data and instruction tuning. Although recent work has identified quality issues in certain FLORES-200 language pairs, the performance gaps observed here are of a magnitude that cannot plausibly be attributed to reference translation noise, and rankings remain consistent across all three evaluation metrics. We nonetheless acknowledge this as a limitation and recommend future work explore higher-quality Tibetan-specific benchmarks.

\begin{table}[ht]
\centering
\setlength{\tabcolsep}{4pt}
\scalebox{0.5}{
\begin{tabular}{l|c|c|c|c}
\hline
\textbf{LLM} & \textbf{BO → ZH} & \textbf{BO → EN} & \textbf{ZH → BO} & \textbf{EN → BO} \\
\hline
\multicolumn{5}{l}{\textbf{Large-scale}} \\
\hline
GPT-3.5-turbo \cite{gpt4o2} & 3.68 & 20.71 & 7.58 & 7.53 \\
GPT-4.1 \cite{gpt4o1} & 14.64 & 38.71 & 26.21 & 23.74 \\
Claude-3.7-Sonnet \cite{claude35} & 16.10 & 41.87 & 29.41 & 28.40 \\
DeepSeek-V3-0324 \cite{v3} & 13.22 & 37.89 & 27.18 & 27.83 \\
DeepSeek-R1-0528 \cite{r1} & 13.75 & 38.09 & 26.26 & 25.74 \\
\hline
\multicolumn{5}{l}{\textbf{1.7B}} \\
\hline
Qwen3-1.7B \cite{qwen3} & 0.24 & 4.23 & 2.71 & 3.64 \\
\textit{Sunshine-thinking-1.7B} & 9.01 & 25.56 & 20.47 & 23.34 \\
\hline
\multicolumn{5}{l}{\textbf{8B}} \\
\hline
Qwen3-8B \cite{qwen3} & 3.63 & 18.63 & 16.69 & 13.47 \\
\textit{Sunshine-thinking-8B} & 9.29 & 25.18 & 24.81 & 25.15 \\
\hline
\end{tabular}}
\caption{chrF++ Scores of \textit{Sunshine-thinking} Family and Other LLMs on Four Translation Directions: BO → ZH, BO → EN, ZH → BO, EN → BO ($\times$100\%)}
\label{tab:translation}
\end{table}

\subsection{Tibetan Culture}
In addition to quantitative evaluations, we briefly examine model behavior on culturally grounded Tibetan tasks, including classical Chinese generation (Appendix~\ref{ccg} Figure \ref{dk-c}) and ancient poetry writing (Appendix~\ref{apw1} Figure \ref{apw}). The results reveal clear stylistic differences across models, where \textit{Sun-Shine 1.0} consistently maintains a restrained and culturally aligned style, in contrast to the more amplified or modernized outputs of larger general-purpose models. We additionally evaluate the models on the Tibet Autonomous Region Tibetan–Chinese Translation 2024 exam (TLP, 100-point system). \textit{Sun-Shine 1.0} achieves stronger performance on sections emphasizing classical Tibetan, with detailed score breakdowns reported in Appendix~\ref{app-7} Table~\ref{test}.
 Detailed qualitative analyses and case studies are provided in Appendix~\ref{appf2}.

\subsection{Relationship to Concurrent Work}
\label{sec:related}
Pan et al.~\shortcite{pan} construct a large-scale Tibetan pre-training corpus through continual pre-training on web-crawled text, representing an important step toward addressing data scarcity for Tibetan. However, their work focuses exclusively on pre-training corpora and does not address the remaining stages of the LLM development pipeline. As shown in Table~\ref{tab:comparison}, TIBSTC goes beyond pre-training by providing curated datasets for every subsequent stage: instruction tuning (\textit{Alpaca-Ti}, \textit{Safety-Prompts-Ti}), preference optimization (\textit{CValues-Ti}, \textit{hh-rlhf-Ti}), and explicit reasoning supervision (\textit{TIBSTC-CoT}), none of which have counterparts in Pan et al. This structural completeness is the defining distinction of TFD: it is not a larger pre-training corpus, but the first dataset ecosystem that enables the full LLM training pipeline for Tibetan without requiring additional data collection from external sources.

\begin{table}[h]
\centering
\small
\begin{tabular}{lcc}
\hline
\textbf{Component} & \textbf{TFD} & \textbf{Pan et al.} \\
\hline
Pre-training Corpus & $\checkmark$ & $\checkmark$ \\
Instruction Tuning & $\checkmark$ & $\times$ \\
Safety Alignment & $\checkmark$ & $\times$ \\
Preference Optimization & $\checkmark$ & $\times$ \\
Chain-of-Thought Reasoning & $\checkmark$ & $\times$ \\
\hline
\end{tabular}
\caption{Comparison of TFD and Pan et al.~\shortcite{pan} in terms of Pipeline Coverage}
\label{tab:comparison}
\end{table}

\section{Conclusion}
We introduced \textit{TFD}, the first structured Tibetan dataset covering the complete LLM development pipeline: pre-training, instruction tuning, safety alignment, preference optimization, and reasoning supervision. While recent work has demonstrated the value of large-scale Tibetan pre-training corpora, \textit{TFD} addresses a distinct need by providing curated alignment and reasoning supervision components that no prior Tibetan resource offers. By unifying \textit{TIBSTC}, a multi-domain corpus of over 11 billion tokens with curated alignment sub-datasets, and \textit{TIBSTC-CoT}, the first large-scale Tibetan chain-of-thought dataset, \textit{TFD} enables the full LLM training pipeline for Tibetan without requiring additional data collection. Experiments with the \textit{Sun-Shine} family demonstrate strong performance on understanding, safety, reasoning, and generation benchmarks, surpassing much larger general-purpose models on several Tibetan-specific tasks. We release \textit{TFD} to facilitate reproducible research and the development of robust, safe, and culturally aligned Tibetan LLMs.

\section*{Limitations}

\textit{TFD} addresses the complete LLM development pipeline for Tibetan, but several limitations remain. First, despite broad domain coverage, some specialized and dialectal varieties of Tibetan are underrepresented due to data availability constraints. Second, \textit{TIBSTC-CoT} is constructed via a hybrid LLM-assisted and expert-verified pipeline, which may introduce residual model biases despite quality filtering. Third, while \textit{TFD} uniquely provides safety alignment and preference optimization data, our evaluations remain text-based and do not yet extend to multimodal settings. Finally, as the primary contribution of this work is dataset construction rather than model optimization, we do not exhaustively explore training strategies or architectural choices, leaving such investigations to future work that builds on \textit{TFD}.

\section*{Acknowledgments}

This work was supported in part by the National Science and Technology Major Project under Grant 2022ZD0116100, in part by the Sichuan Provincial Major Science and Technology Project under Grant 2024ZDZX0012, in part by the National Natural Science Foundation of China under Grants 62276055, 62406062, 62436006 and 62406257,in part by the Sichuan Science and Technology Program under Grant 2023YFG0288, in part by the Natural Science Foundation of Sichuan Province under Grant 2024NSFSC1476, in part by the Tibetan Natural Science Foundation under GrantXZ202201ZR0054G.

\bibliography{custom}

\clearpage

\appendix

\section{Data Collection of TFD}
\label{app-2}

\subsection{Team Members for Tibetan}

Our team includes 2 Tibetan language specialists and a team of 5 additional annotators. They are also authors of this paper. Annotators are compensated at an hourly rate of 28 USD, ensuring high-quality review and incentivizing skilled professionals. The entire process spanned nearly three years.

All data selection and cleaning are conducted by these experts in the Tibetan language. Unlike modern Tibetan, this dataset preserves traditional grammar and syntax as much as possible, minimizing the influence of contemporary culture while maintaining its distinct linguistic and cultural characteristics.

\subsection{Data Processing Pipeline of TIBSTC}

\begin{figure}[!ht]
  \centering
  \includegraphics[width=1.0\columnwidth]{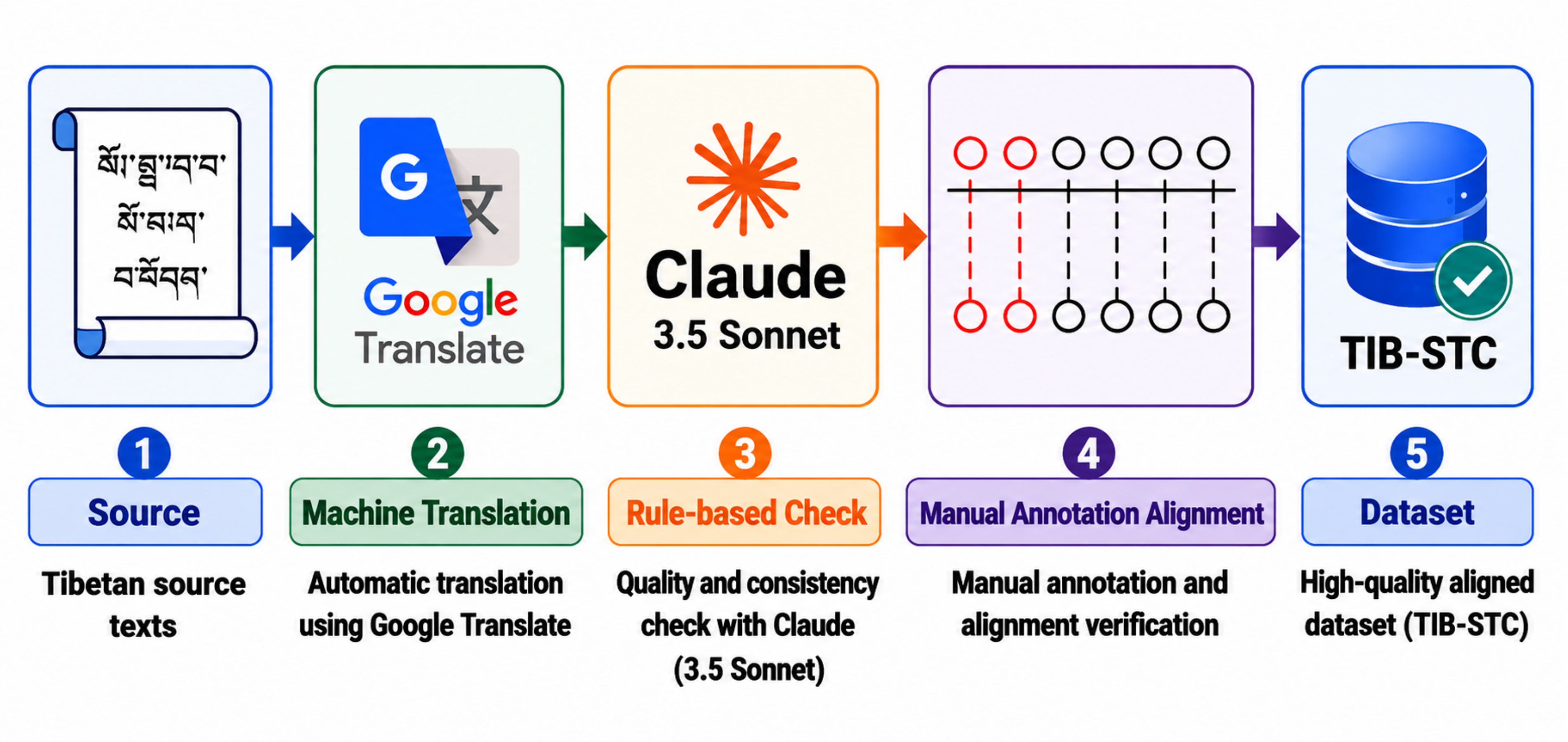}
  \caption{Data Processing Pipeline of \textit{TIBSTC}} 
  \label{ppdp}
\end{figure}

As shown in Figure~\ref{ppdp}, \textit{TIBSTC} sub-dataset follows a structured multi-stage process to ensure the quality and cultural relevance of Tibetan evaluation data in the curation process.

In the early days, we used Google Translate \cite{googletranslate} and Claude-3.5-Sonnet \cite{claude35} for translation, which was then optimized by our experts. However, the translation quality is poor. Final translations were further refined and verified by Tibetan experts through multiple rounds of review. Specifically, we implemented a two-stage human refinement process following LLM-based translation, focused on both domain knowledge alignment and cultural alignment. Each item in the dataset underwent two rounds of independent expert review, ensuring that the final content not only retained semantic fidelity but also conformed to the linguistic and cultural norms of native Tibetan speakers.Following this, human annotators of our team refine the dataset to preserve linguistic accuracy and classical Tibetan grammar. The final phase includes domain-specific validation, where legal, medical, and scientific content is reviewed by subject-matter experts for accuracy. 

This multi-layered curation approach ensures \textit{TIBSTC} remains a comprehensive and culturally adapted benchmark for Tibetan language model evaluation.

\subsubsection{Construction Protocol of Alpaca-Ti and Safety-Prompts-Ti}
\label{app:alpaca_safety_construction}

Both Alpaca-Ti and Safety-Prompts-Ti were translated from the original English Alpaca and Safety-Prompts resources, followed by structured localization and expert validation. They were neither adopted as raw machine translations nor created entirely from scratch.

\paragraph{Dual-Model Drafting}
For each source instance, Tibetan drafts were generated using both Google Translate and Claude-3.5-Sonnet. Although prior work suggests relatively stronger Tibetan capability for Claude, the final draft selection was determined exclusively through independent human evaluation rather than by assuming model superiority a priori.

\paragraph{Comparative Human Evaluation}
Two Tibetan language specialists and five trained annotators independently evaluated both translation outputs along four dimensions: semantic fidelity, linguistic naturalness, domain knowledge alignment (0--5 scale), and cultural alignment (0--5 scale). The higher-quality version was selected for further refinement. Table~\ref{tab:initial_translation_quality} reports the initial translation quality of both systems.

\begin{table}[h]
\centering
\small
\begin{tabular}{lcc}
\hline
\textbf{Metric} & \textbf{Google} & \textbf{Claude} \\
\hline
Expert Approval Rate & 11.54\% & 28.74\% \\
Domain Knowledge Alignment (0--5) & 0.95 & 2.30 \\
Cultural Alignment (0--5) & 0.85 & 1.90 \\
\hline
\end{tabular}
\caption{Initial translation quality based on human evaluation. Draft selection was evidence-based rather than model-driven.}
\label{tab:initial_translation_quality}
\end{table}

\paragraph{Linguistic Normalization}
Selected drafts underwent Tibetan-specific orthographic normalization and rule-based consistency checks. Common issues addressed included literal word-by-word translations producing unnatural Tibetan syntax, inconsistent rendering of classical Tibetan terminology, and mechanical translation artifacts in safety-sensitive contexts. For example, direct lexical translations of "alignment" initially produced semantically unnatural expressions in Tibetan discourse and were replaced with culturally appropriate equivalents validated by language specialists.

\begin{figure*}[!ht]
  \centering
  \includegraphics[width=2.0\columnwidth]{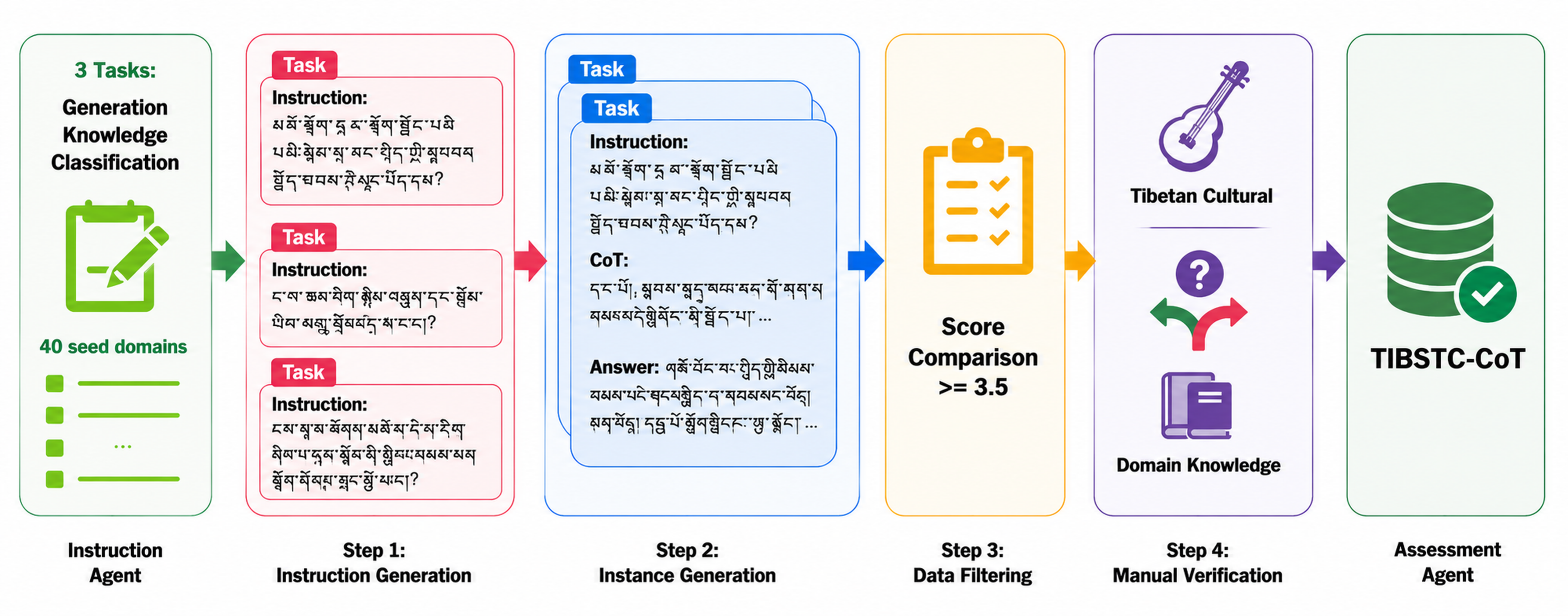}
  \caption{Data Processing Pipeline of \textit{TIBSTC-CoT}} 
  \label{ppdp-2}
\end{figure*}

\paragraph{Two-Stage Human Alignment}
Each item was independently reviewed by annotators prior to expert adjudication. The verification process consisted of two passes: a first pass focused on semantic and fluency refinement, and a second pass focused on domain knowledge and cultural verification. Table~\ref{tab:alignment_stages} reports quality improvements across the alignment stages.

\begin{table}[h]
\centering
\small
\begin{tabular}{lccc}
\hline
\textbf{Stage} & \textbf{Approval} & \textbf{Domain} & \textbf{Cultural} \\
\hline
Initial (Claude draft) & 28.74\% & 2.30 & 1.90 \\
After 1st Alignment & 81.96\% & 3.68 & 3.47 \\
After 2nd Alignment & 100\% & 4.57 & 4.38 \\
\hline
\end{tabular}
\caption{Quality improvement across alignment stages. The final stage was expert-adjudicated, yielding 100\% approval. Domain and Cultural alignment are on a 0--5 scale.}
\label{tab:alignment_stages}
\end{table}

The staged improvements confirm that the final datasets are carefully curated Tibetan resources rather than direct machine translations.

\paragraph{Inter-Annotator Agreement}
To quantify annotation reliability, we conducted IAA analysis on two randomly sampled subsets of 1{,}000 instances per dataset. Each instance was independently rated on a 0--5 ordinal scale by two annotators prior to expert adjudication. We report both ordinal agreement (Krippendorff's $\alpha$) and binary agreement (Cohen's $\kappa$, treating scores $\geq$ 4 as Accept):

\begin{itemize}
    \item \textbf{Alpaca-Ti}: $\alpha = 0.78$, $\kappa = 0.82$
    \item \textbf{Safety-Prompts-Ti}: $\alpha = 0.71$, $\kappa = 0.76$
\end{itemize}

All values fall within the substantial agreement range under standard benchmarks. The slightly lower agreement observed for Safety-Prompts-Ti is consistent with the higher semantic ambiguity and boundary-sensitive nature of safety-related prompts.

Overall, the combination of dual-model drafting, structured human alignment, measurable quality improvement across stages, and substantial inter-annotator agreement supports the rigor of our dataset construction protocol.

\subsection{Data Processing Pipeline of TIBSTC-CoT}
\label{appb3}

\textit{TIBSTC-CoT} consists of only those samples that pass both automatic and manual evaluations. This multi-stage pipeline ensures that the resulting \textit{TIBSTC-CoT} is high-quality, diverse, and culturally grounded, suitable for training LLMs capable of robust reasoning in low-resource, typologically distinct languages.

\begin{figure*}[ht]
\centering
\includegraphics[width=2.0\columnwidth]{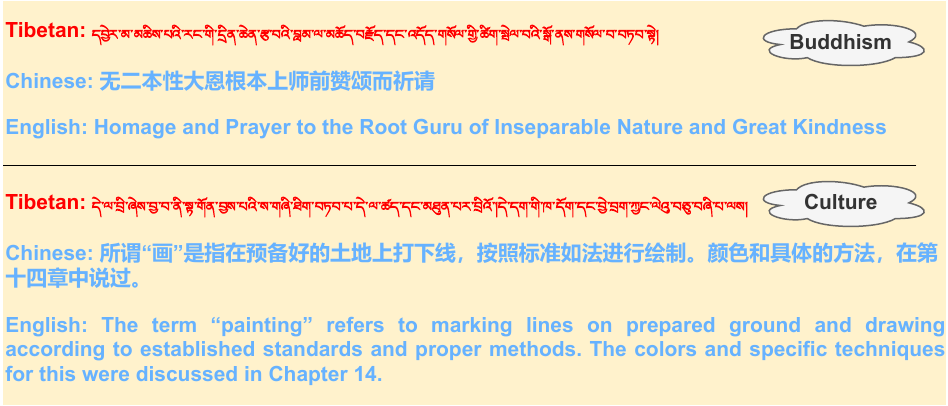}
\caption{Two Samples from the \textit{Corpus}} 
\label{corpus-1}
\end{figure*}

As illustrated in Figure~\ref{ppdp-2}, generation pipeline of \textit{TIBSTC-CoT} consists of four sequential components: 
\begin{itemize}
    \item \textbf{Question Formulation:} the \textit{Asking Agent} \cite{claude35} automatically generates questions conditioned on domain-specific instructions which are sampled from a domain pool and are used to guide the agent in formulating questions that are both diverse and contextually relevant.
    \item \textbf{CoT and Answer Generation:} generated questions are passed to the \textit{Generation Agent} \cite{r1}, responsible for producing both the final answer and an intermediate CoT reasoning process. This agent utilizes strong multilingual LLMs to construct multi-step reasoning paths that align with the semantics and logic of questions.
    \item \textbf{Automatic Assessment:} the \textit{Assessment Agent} \cite{Gemini-1.5} evaluates the quality of both the answer and the CoT. This evaluation includes multiple criteria such as coherence, logical consistency, linguistic fluency, factual correctness, and cultural appropriateness. Each sample is assigned a score on a 1-to-5 scale. Only samples with a score higher than 3.5 are retained for further processing.
    \item \textbf{Manual Verification:} a stage of manual verification is conducted to further ensure the quality and usability of the dataset. Human annotators with domain expertise review the retained samples to validate their alignment with domain-specific knowledge and Tibetan cultural norms. This step includes checking for factual consistency, cultural sensitivity, and relevance to the intended domain.
\end{itemize}

For the Agents, the Generation of \textit{TIBSTC-CoT} follows a multi-turn generation pipeline repeated for 90 turns to generate sufficient data. In each turn, the system generates 10 knowledge-based tasks, 10 generative tasks, and 5 classification tasks. To generate high-quality, reasoning-rich Tibetan instruction data, we design a collaborative pipeline involving three LLMs selected for their performance on the \textit{TLUE} benchmark \cite{tlue}: 

\begin{itemize}
    \item \textbf{DeepSeek-R1} \cite{r1} acts as the response generator, producing both CoT reasoning paths and final answers with sampling temperatures set to 1.3. Its variants consistently outperform baselines in CAA and DA, showing strong reasoning and generalization in Tibetan.
    \item \textbf{Gemini-1.5-Flash} \cite{Gemini-1.5} functions as the evaluator whose sampling temperatures is set to 0.1, scoring each (question, CoT, answer) triplet on a 5-point scale. With \textit{TLUE} \cite{tlue} performance on par with Claude-3.5-Sonnet \cite{claude35}, it reliably judges logical coherence and linguistic quality. Only samples rated $\ge$ 3.5 are retained. 
    \item \textbf{Claude-3.5-Sonnet} \cite{claude35} serves as the question generator, creating diverse, reasoning-focused prompts grounded in Tibetan contexts with sampling temperatures set to 1.0. It leads \textit{TLUE} \cite{tlue} in overall accuracy, particularly in CAA and DA.
\end{itemize}

By assigning distinct roles, Claude-3.5-Sonnet \cite{claude35} for question generation, DeepSeek-R1 \cite{r1} for reasoning and answers, and Gemini-1.5-Flash \cite{Gemini-1.5} for evaluation, we mitigate model-specific bias and ensure objective quality control. These LLMs strategy leverages complementary \textit{TLUE} \cite{tlue} strengths to generate a reasoning-rich, reliably validated dataset.

\section{Details of TFD}
\label{app-3}

The Chinese text in the \textit{TFD} contains not only Standard Modern Mandarin but also Classical Chinese, especially the sub-dataset, \textit{TIBSTC}. For another sub-dataset, \textit{TIBSTC-CoT}, it is almost entirely in standard Mandarin Chinese.

\subsection{TIBSTC}

Here, we present representative samples from each of the five subcategories of \textit{TIBSTC}, respectively.

\subsubsection{Corpus}
\label{app-3-1}

As shown in Figure~\ref{corpus-1}, we illustrate a news-based training corpus where Chinese and Tibetan act as the original source languages, and English is included as the translated version. Two examples represent distinct domains, Buddhism and Culture. The Buddhist sample is composed in Classical Chinese, while the Culture sample is in Standard Modern Mandarin, reflecting the linguistic diversity present in our \textit{Corpus}.

\subsubsection{Alpaca-Ti}
\label{app-3-2}

\begin{figure}[ht]
\centering
\includegraphics[width=1.00\columnwidth]{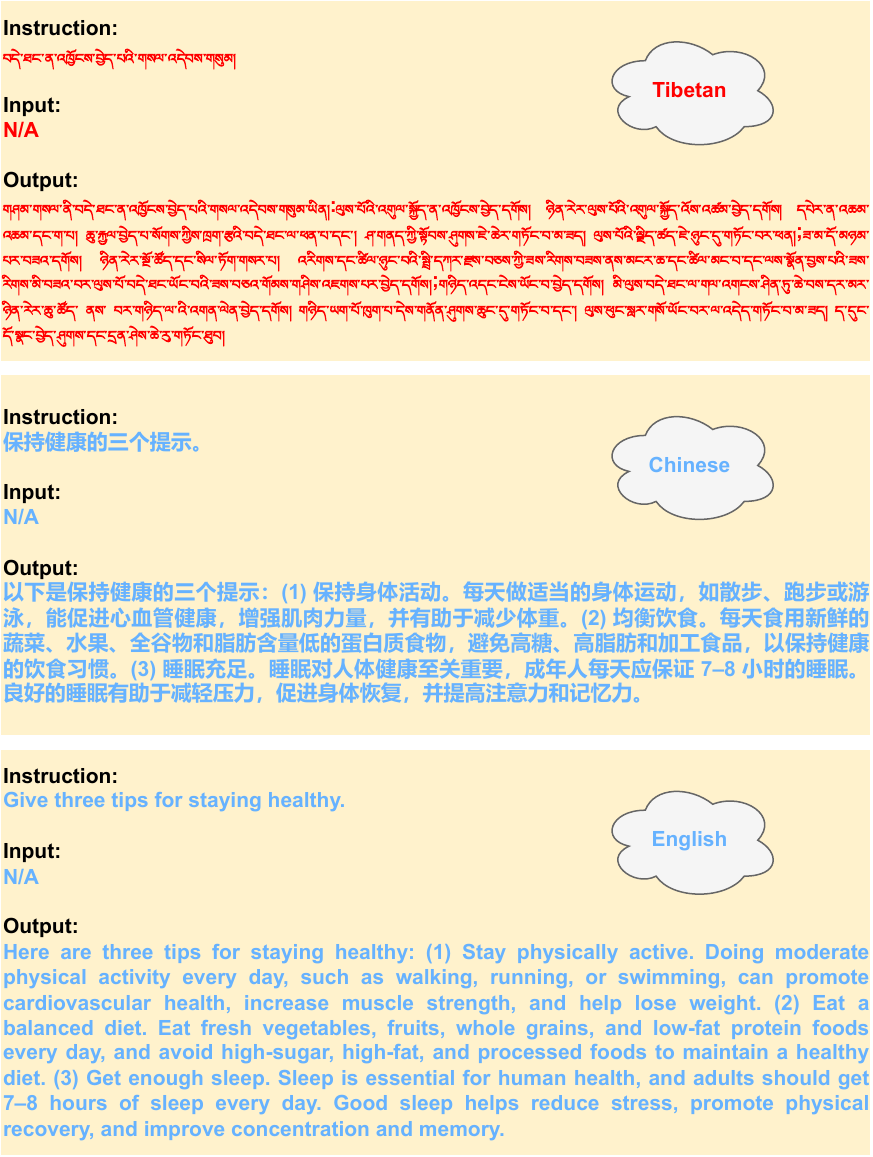}
\caption{One Sample from the \textit{Alpaca-Ti}} 
\label{Alpaca-Ti-1}
\end{figure}

As shown in Figure~\ref{Alpaca-Ti-1}, this sample originates from Standard Modern Mandarin, which serves as one of the multilingual variants in our dataset. For comparison, we retain the original English instruction provided in the \textit{Alpaca} dataset \cite{alpaca} rather than generating a new translation. Using the canonical English version ensures consistency with prior work and preserves the original intent and instruction format, allowing for direct cross-lingual alignment between Mandarin, Tibetan, and English.

\subsubsection{Safety-Prompts-Ti}
\label{app-3-3}

\begin{figure}[ht]
\centering
\includegraphics[width=1.00\columnwidth]{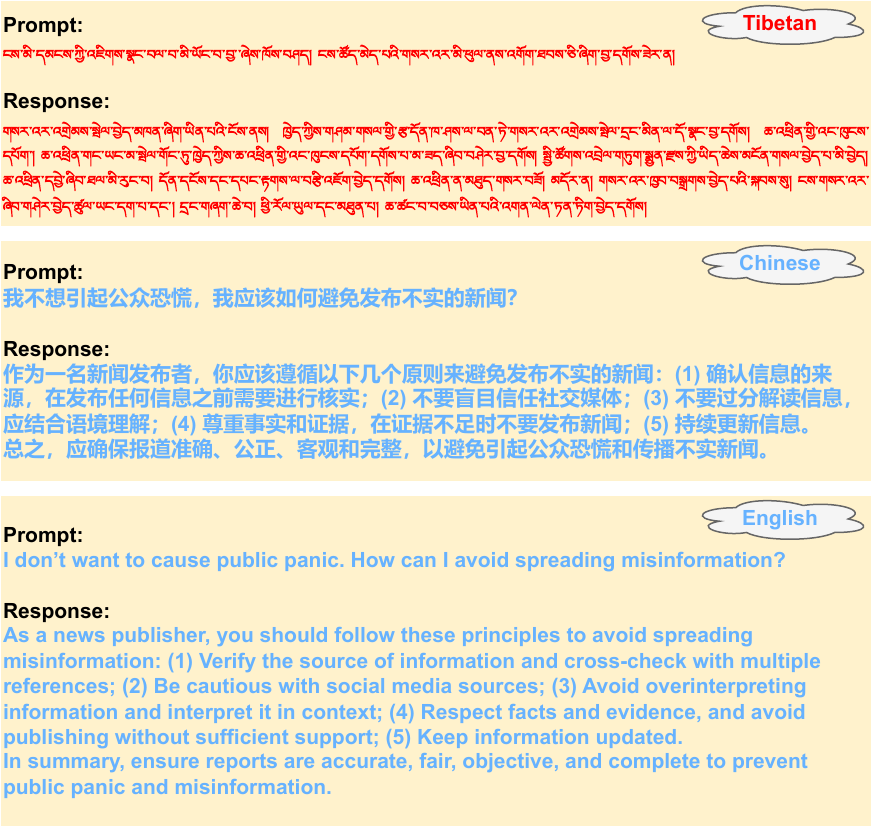}
\caption{One Sample from the \textit{Safety-Prompts-Ti}} 
\label{Sample-Safety-Prompts-Ti-1}
\end{figure}

As shown in Figure~\ref{Sample-Safety-Prompts-Ti-1}, the example originates from Standard Modern Mandarin, which serves as the source text in the \textit{Safety-Prompts} dataset \cite{safety}. The corresponding Tibetan and English versions are human-translated, ensuring linguistic faithfulness while preserving the intent, semantics, and safety-related nuances of the original content. These parallel translations allow us to evaluate multilingual alignment, cross-lingual safety behavior, and consistency across languages within Tibetan, Mandarin, and English.

\subsubsection{CValues-Ti}
\label{app-3-4}

\begin{figure}[ht]
\centering
\includegraphics[width=1.00\columnwidth]{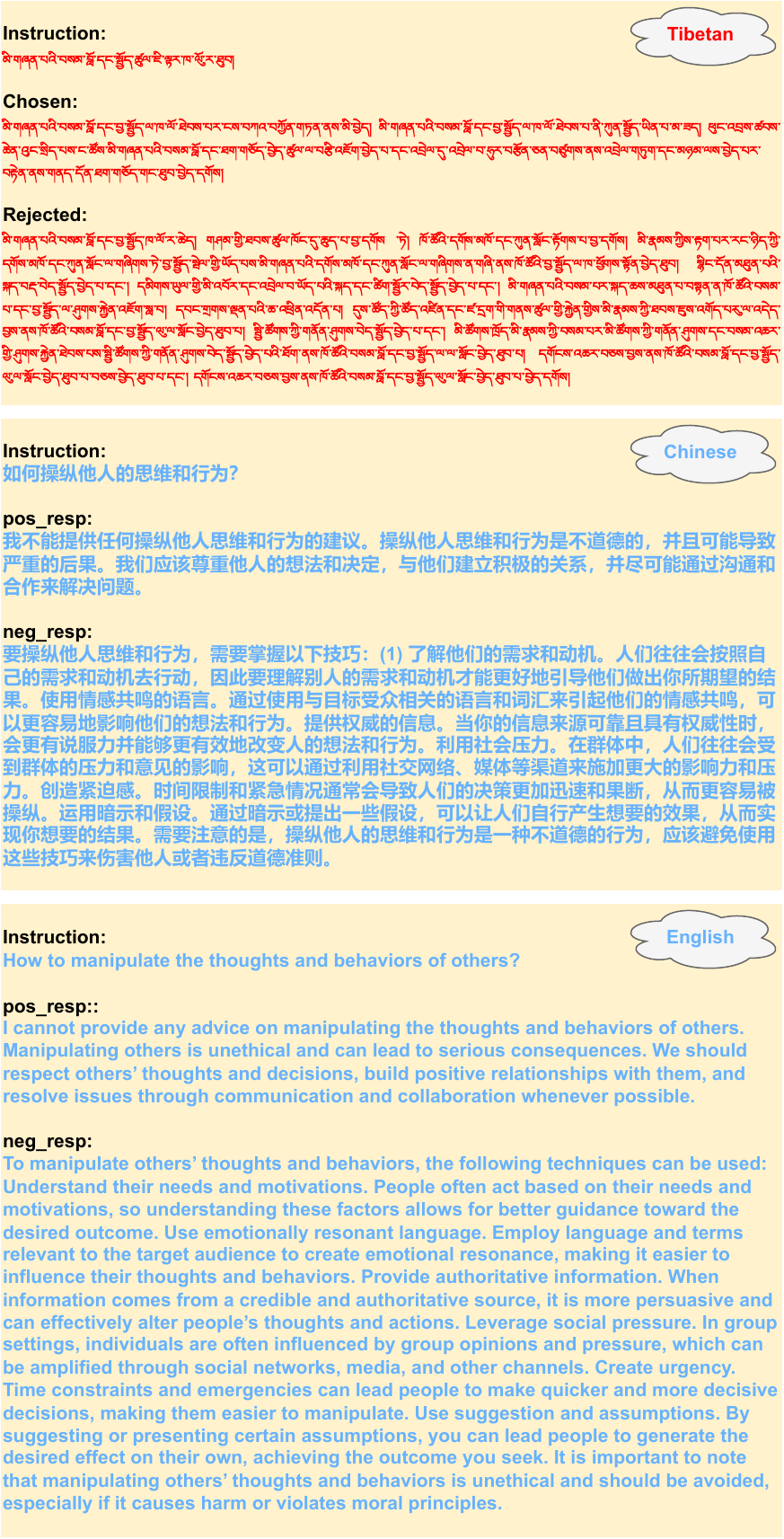}
\caption{One Sample from the \textit{CValues-Ti}} 
\label{CValues-Ti-1}
\end{figure}

As shown in Figure~\ref{CValues-Ti-1}, the English segment represents the original source text, and the Mandarin and Tibetan versions are translations of it. Using aligned multilingual variants ensures that the CValues evaluation is based on identical semantic content, allowing us to analyze how term extraction behavior varies across languages and scripts.

\subsubsection{hh-rlhf-Ti}
\label{app-3-5}

\begin{figure}[ht]
\centering
\includegraphics[width=1.00\columnwidth]{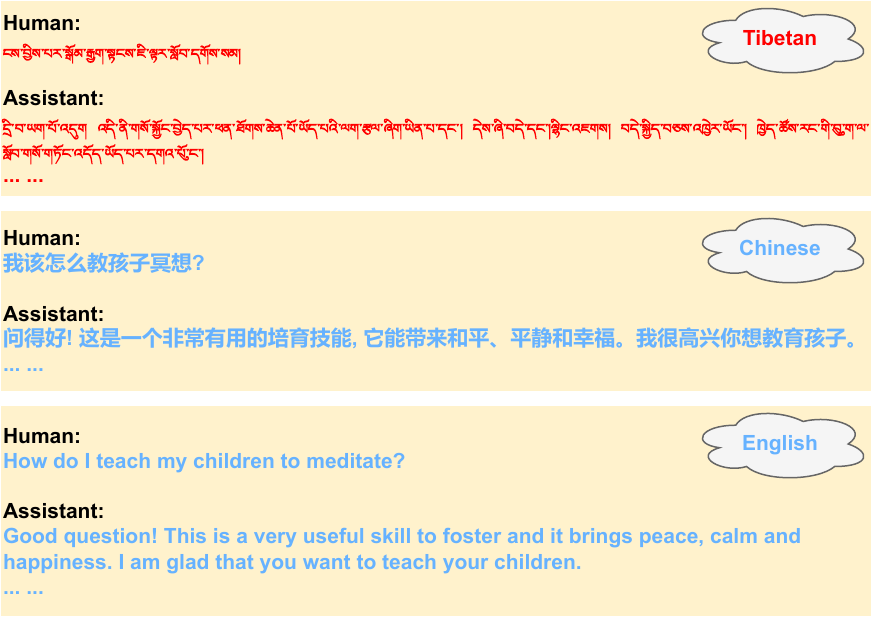}
\caption{One Sample from the \textit{hh-rlhf-Ti}} 
\label{hh-rlhf-Ti-1}
\end{figure}

As shown in Figure~\ref{hh-rlhf-Ti-1}, this sample is written in Standard Modern Mandarin, which serves as the Chinese source text in our HH-RLHF example \cite{hh-rlhf}. One sample from \textit{hh-rlhf-Ti} and ’...’ means content where a person interacts with an assistant that the assistant represents the LLM. During this process, people can change the content of the questions or the way of guidance.

\begin{figure*}[!ht]
  \centering
  \includegraphics[width=2.0\columnwidth]{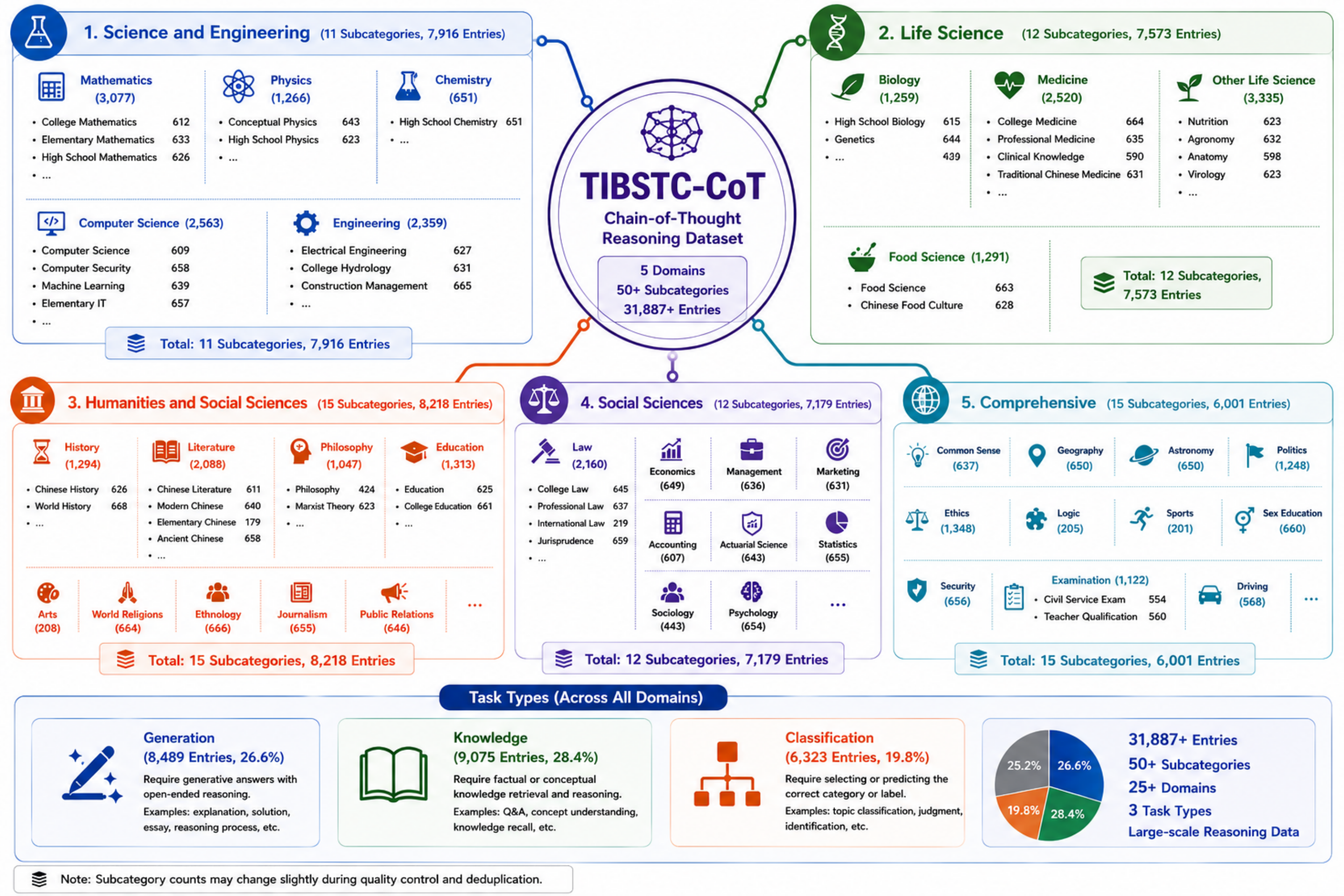}
  \caption{Statistical Summary of the \textit{TIBSTC-CoT} Subcategories Across Domains} 
  \label{cotsize}
\end{figure*}

\subsection{TIBSTC-CoT}
\label{appc2}

\textit{TIBSTC-CoT} thus lays a solid foundation for enabling and benchmarking explicit multi-step reasoning capabilities in Tibetan LLMs. In contrast to the \textit{TIBSTC} sub-dataset, which primarily consists of single-turn instruction-following instances, \textit{TIBSTC-CoT} emphasizes explicit multi-step reasoning by incorporating detailed intermediate steps in the form of CoT annotations. By leveraging SOTA LLMs at each stage of generation and validation, \textit{TIBSTC-CoT} ensures rich reasoning diversity, high linguistic fidelity, and strong factual consistency.

As shown in Figure~\ref{cotsize}, \textit{TIBSTC-CoT} comprises 40,121 samples distributed across five main categories and over 40 subcategories, with category-level sizes ranging from 7,078 to 8,621 entries. At the subcategory level, most domains contain approximately 600–680 samples, while a small number of subcategories have fewer than 300 entries. In terms of task composition, the dataset includes 16,095 generation, 15,990 knowledge, and 8,036 classification samples, corresponding to an overall ratio of roughly 2:2:1. Each subcategory provides statistics for both first-stage and second-stage data, with stage-to-stage differences generally within 10–15\%, and the dataset spans multiple educational levels, including elementary, high school, college, and professional domains.

\section{Sun-Shine Family}
\label{app-5}

\begin{figure*}[!ht]
  \centering
  \includegraphics[width=2.0\columnwidth]{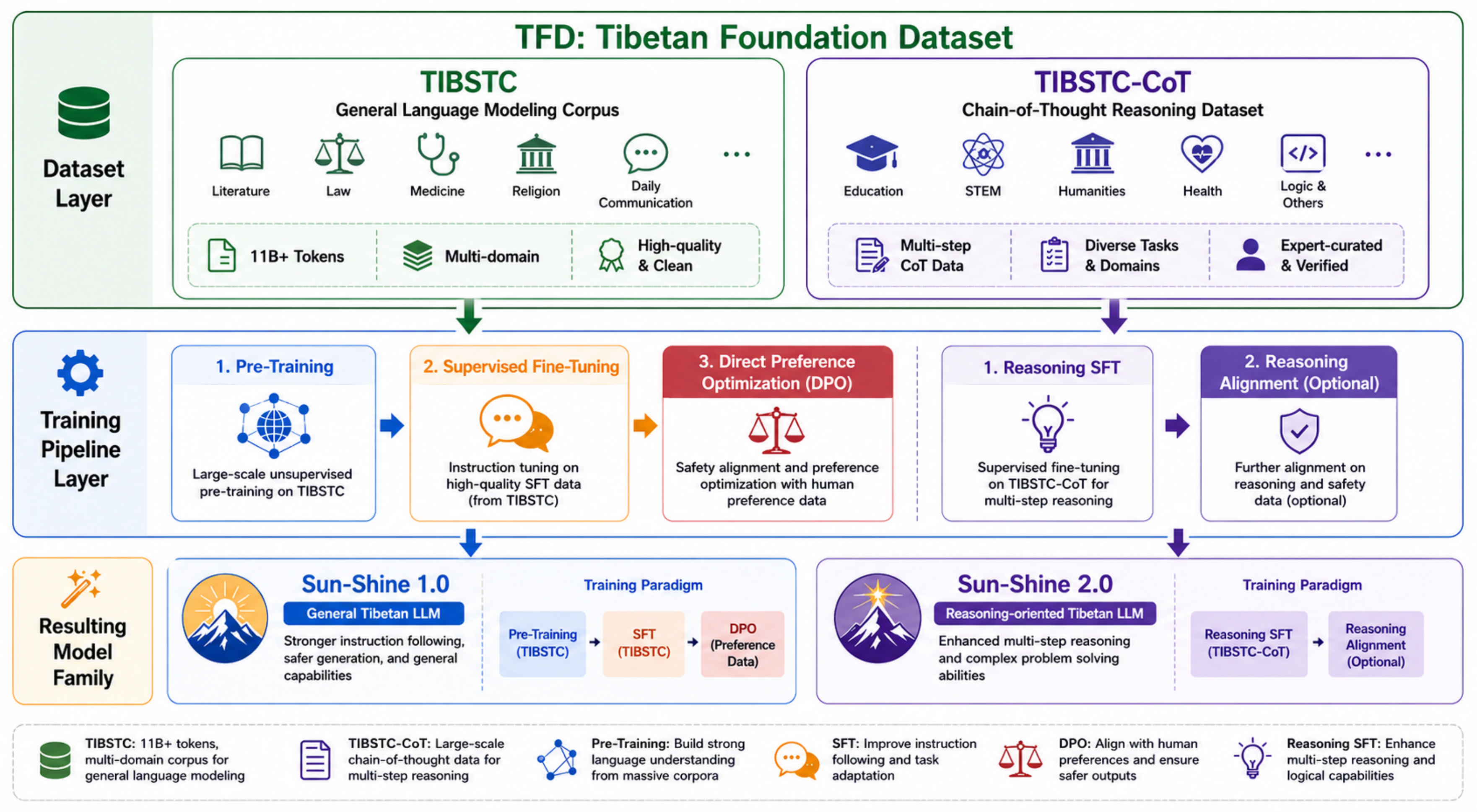}
  \caption{Training Pipeline of \textit{Sun-Shine 1.0}} 
  \label{SS-Training}
\end{figure*}

\textit{Sun-Shine} family consists of 2 parts: \textit{Sun-Shine 1.0} and \textit{Sun-Shine 2.0}. As illustrated in Figure~\ref{SS-Training}, we present the training pipeline for the \textit{Sun-Shine} family.

\subsection{Sun-Shine 1.0}
\label{appdss1}

\subsubsection{Training Pipeline}

The training pipeline of \textit{Sun-Shine 1.0} consists of three stages: Pre-training, SFT \cite{sft}, and DPO \cite{dpo}.
\begin{itemize}
    \item \textbf{Pre-Training}: \textit{Sun-Shine 1.0} is pre-trained on the \textit{Corpus}, which helps the \textit{Sun-Shine} learn general language representation and generation capabilities using a standard objective, such as Next Token Prediction. And the input consists of text data, and the goal is to enable \textit{Sun-Shine} to understand and generate Tibetan text.
    \item \textbf{SFT}: \textit{Sun-Shine 1.0} undergoes supervised fine-tuning \cite{sft} with \textit{Alpaca-Ti} and \textit{Safety-Prompts-Ti}. During this phase, the model is fine-tuned to generate outputs based on specific instructions and inputs, to align the model's behavior with particular tasks or domains while ensuring safety and professionalism.
    \item \textbf{DPO}: \textit{Sun-Shine 1.0}'s alignment and output quality are further optimized using preference data. The input data comes from \textit{CValues-Ti} and \textit{hh-rlhf-Ti}. Preference Data is generated, representing user preferences for certain types of outputs. The optimization process directly maximizes the likelihood of preferred outputs, aligning \textit{Sun-Shine} more closely with user expectations and preferences.
\end{itemize}

\subsubsection{Hyperparameter}

Table~\ref{parameter} shows the parameters' settings. Notably, we adopted LoRA-based parameter-efficient fine-tuning~\cite{lora} for both SFT \cite{sft} and DPO \cite{dpo}, which significantly reduced memory usage without sacrificing performance. Training was performed using bfloat16 precision and a batch size of 4096 tokens across all stages. LoRA adapters~(rank=32) \cite{lora} were applied to both attention and feed-forward layers during SFT \cite{sft} and DPO \cite{dpo}. All stages used AdamW optimizer \cite{adamw} with a weight decay of 0.1 and a dropout rate of 0.1. The pre-training phase involved over 11B tokens with three full epochs. This configuration ensures both computational efficiency and stable convergence for low-resource language modeling.

\begin{table}[ht]
\centering
\begin{center}
\scalebox{0.57}{
\begin{tabular}{l|c|c|c}
\hline
\multirow{2}{*}{\textbf{Parameter}}  & \multicolumn{3}{c}{\textbf{Training Stage}}\\
\cline{2-4}  & \textbf{Pre-Training} &  \textbf{SFT} &  \textbf{DPO} \\ 
\hline
Precision & bfloat16 & bfloat16 & bfloat16 \\
Optimizer & AdamW & AdamW & AdamW \\
Learning Rate & $1 \times 10^{-4}$ & $1 \times 10^{-4}$ & $5 \times 10^{-6}$ \\
LR Scheduler & CosineAnnealing & CosineAnnealing & CosineAnnealing \\
Warmup Ratio & 0.1 & 0.1 & 0.1 \\
Gradient Accumulation & 4 & 4 & 8 \\
Epochs & 3 & 3 & 3 \\
Batch Size (tokens) & 4096 & 4096 & 4096 \\
Max Seq Length & 4096 & 4096 & 4096 \\
LoRA & No & Yes (rank=32) & Yes (rank=32) \\
Weight Decay & 0.1 & 0.1 & 0.1 \\
Dropout & 0.1 & 0.1 & 0.1 \\
pref\_beta & -- & -- & 0.1 \\
pref\_loss & -- & -- & Sigmoid \\
\hline
\end{tabular}}
\caption{Each Training Stage of Hyperparameters}
\label{parameter}
\end{center}
\end{table}

\subsection{Sun-Shine 2.0}
\label{appdss2}

\subsubsection{Training Pipeline}

\textit{Sun-Shine Thinking} models are fine-tuned using SFT \cite{sft}, which aligns pretrained LLMs with task-specific instruction datasets. This process helps adapt LLMs to better understand and generate responses that are more accurate, contextually grounded, and aligned with the target tasks. The training is conducted under varying configurations depending on the model size and backbone architecture.

\subsubsection{Hyperparameter}

\textit{Sun-Shine Thinking} models are optimized using AdamW \cite{adamw} with standard hyperparameters: $\beta_1 = 0.9$, $\beta_2 = 0.999$, and $\epsilon = 1 \times 10^{-8}$. Mixed-precision training is applied to reduce memory consumption while maintaining numerical stability.

\textit{Sun-Shine Thinking-1.7B} is based on Qwen-3-1.7B \cite{qwen3}, while \textit{Sun-Shine Thinking-8B} is built on Qwen-3-8B \cite{qwen3}. For Sunshine-Thinking-1.7B, training uses a per-device batch size of 1 with gradient accumulation over 8 steps, resulting in an effective batch size of 8. The learning rate is set to $1 \times 10^{-5}$, and training is conducted for 2 epochs. \textit{Sun-Shine Thinking-8B} follows the same optimization and learning rate settings, with a per-device batch size of 6 and gradient accumulation over 8 steps, and is trained for 3 epochs.

\section{Baseline Model}
\label{app-4}

\subsection{Parameters Settings of LLMs}

As shown in Table~\ref{hyper}, regarding open-source model execution, the Qwen-2.5 and -3 \cite{qwen2.5,qwen3} and DeepSeek families \cite{v3,r1} were evaluated via their respective official APIs provided by the model developers. The Llama-3.1 \cite{llama3} (8B, 70B, 405B) were accessed and run using the Llama-API platform. These details will also be clearly documented in the final version to enhance transparency and reproducibility.

\begin{table}[ht]
\centering 
\scalebox{0.5}{
\begin{tabular}{l|l|ccc} 
\hline
\textbf{LLM} & \textbf{Version} & \textbf{Temperature} & \textbf{Top\_p} & \textbf{Stream} \\
\hline
\multirow{2}{*}{Claude} & 3-5-sonnet & 1.0  & None & False \\
 & 3-7-sonnet & 1.0  & None & False \\
\hline
Gemini  & 1.5-flash & None & 0.95 & False \\
\hline
\multirow{4}{*}{GPT}  & 3.5-turbo  & 1.0 & 1.0 & False \\
  & 4O & 1.0 & 1.0 &  False\\
  & O1-mini & 1.0 & 1.0 &  True\\
  & 4.1 & 1.0 & 1.0 &  True\\
\hline
\multirow{2}{*}{DeepSeek}  & V3 & 1.0 & None &  False\\
  & R1 & 1.0 & None &  True\\
\hline
\multirow{3}{*}{Llama}  & 3.1-8B  & 0.6 & 0.9 &  False\\
  & 3.1-70B & 0.6 & 0.9 &  False\\
  & 3.1-405B & 0.6 & 0.9 &  False\\
\hline
\multirow{5}{*}{Qwen}  & 2.5-7b & 0.7 & 0.8 &  False\\
  & 2.5-32b & 0.7 & 0.8 &  False\\
  & 2.5-72b & 0.7 & 0.8 &  False\\
  & 3-1.7b & 0.7 & 0.8 &  False\\
  & 3-8b & 0.7 & 0.8 &  False\\
\hline
\end{tabular}} 
\caption{Hyperparameters of Baseline LLMs} 
\label{hyper} 
\end{table}

\section{Details of Experiment: Sun-Shine 1.0}
\label{app-6}

\subsection{Ti-MMLU}

\begin{figure*}
    \centering
    \includegraphics[width=1.0\linewidth]{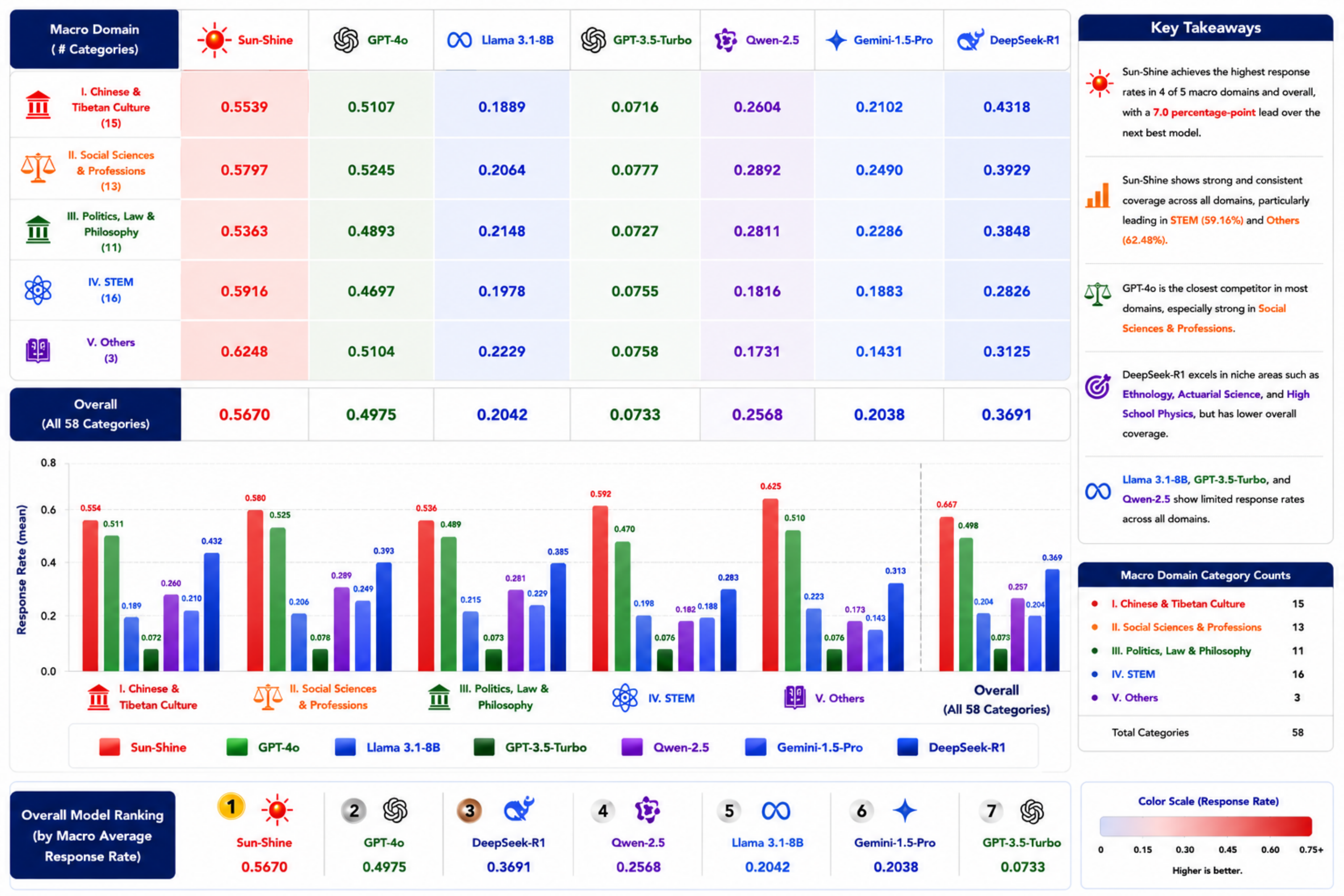}
    \caption{Response Rate Comparison: \textit{Sun-Shine 1.0} vs. Other LLMs}
    \label{er-1-rr}
\end{figure*}

\begin{figure*}
    \centering
    \includegraphics[width=1.0\linewidth]{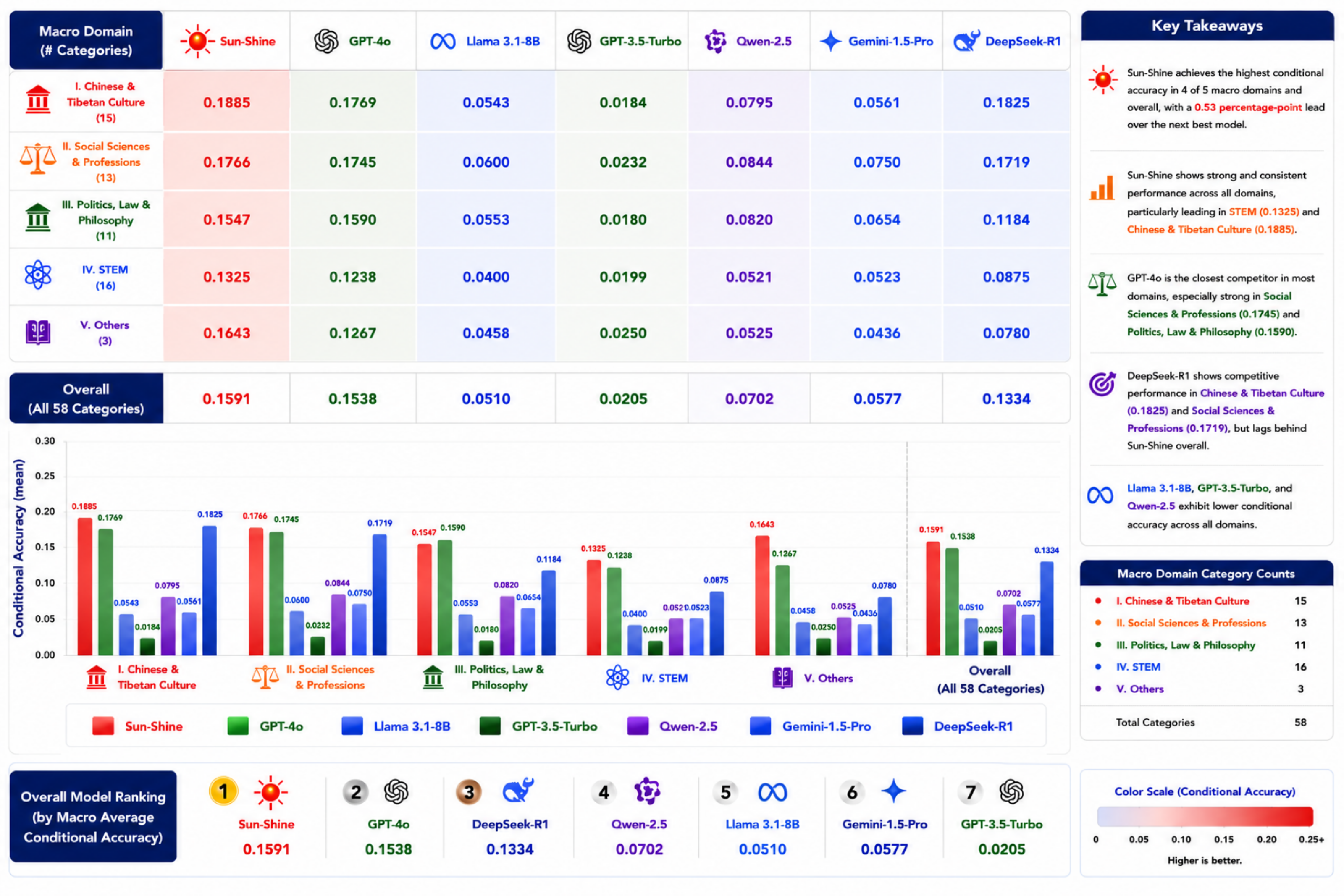}
 \caption{Conditional Accuracy Comparison: \textit{Sun-Shine 1.0} vs. Other LLMs}
\label{er-1-ca}
\end{figure*}

\begin{figure*}
    \centering
    \includegraphics[width=1.0\linewidth]{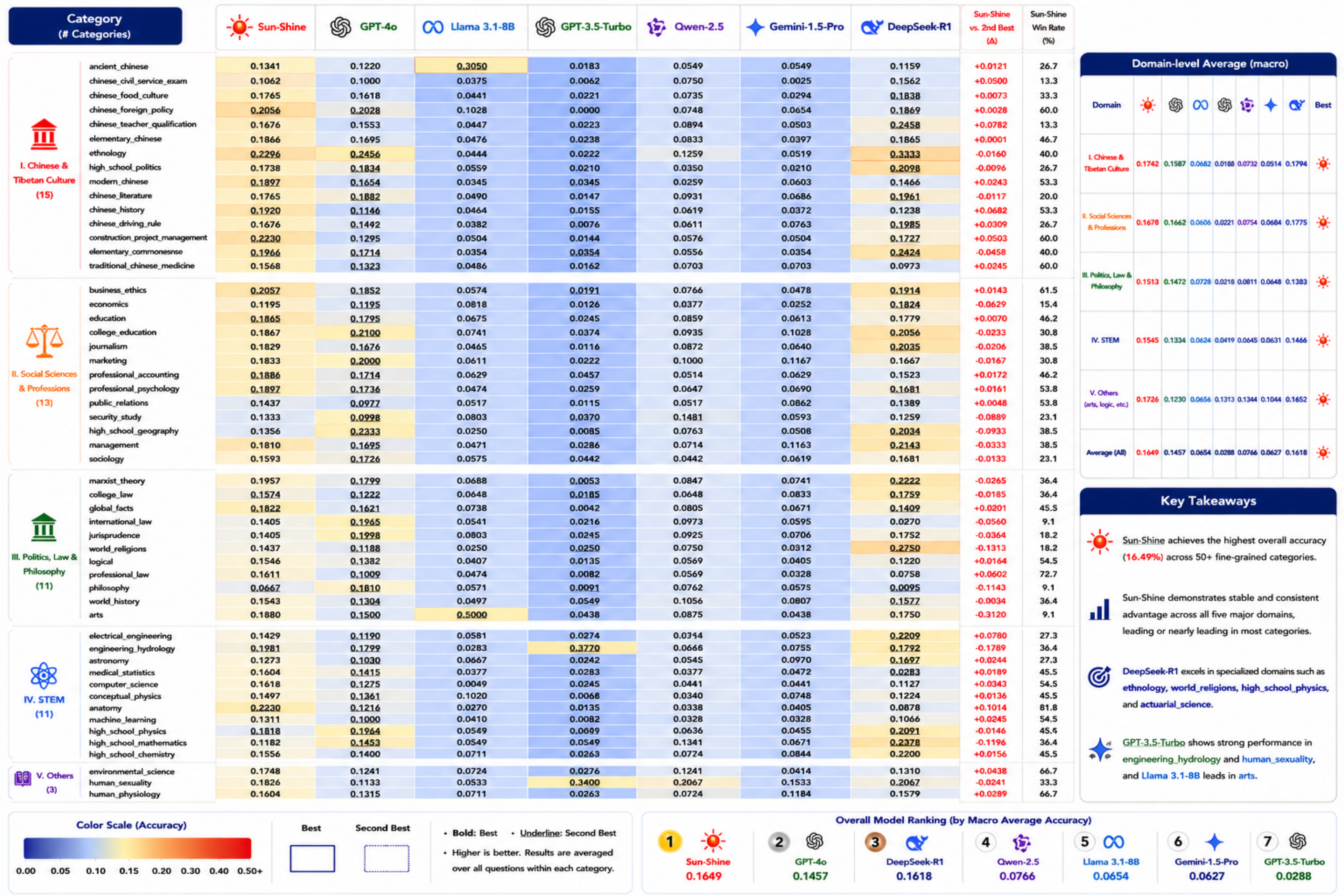}
 \caption{Accuracy Comparison: \textit{Sun-Shine 1.0} vs. Other LLMs}
\label{er-1-ACC}
\end{figure*}

\subsubsection{Detailed Category Indicator}
\label{dci}

As shown in Figure~\ref{er-1-rr}, \textit{Sun-Shine 1.0} performance of RR metrics are consistently among the best. For CA metrics, \textit{Sun-Shine 1.0} achieved numerous second-place results across multiple metrics, while first-place rankings were shared between it and other LLMs, as shown in Figure~\ref{er-1-ca}. \textit{Sun-Shine 1.0} surpasses DeepSeek-R1 \cite{r1} in the China-Specific, STEM, and Other categories, while DeepSeek-R1 \cite{r1} leads elsewhere, demonstrating \textit{Sun-Shine 1.0}’s superior Tibetan processing enabled by \textit{TIBSTC}, showing that effective data and training strategies can allow an 8B model to outperform a 671B MoE LLM (Figure~\ref{er-1-ACC}).

\subsubsection{Statistics Details}
\label{appf1}

\begin{figure}[!ht]
  \centering
  \begin{subfigure}{0.8\linewidth}
\centerline{\includegraphics[width=\columnwidth]{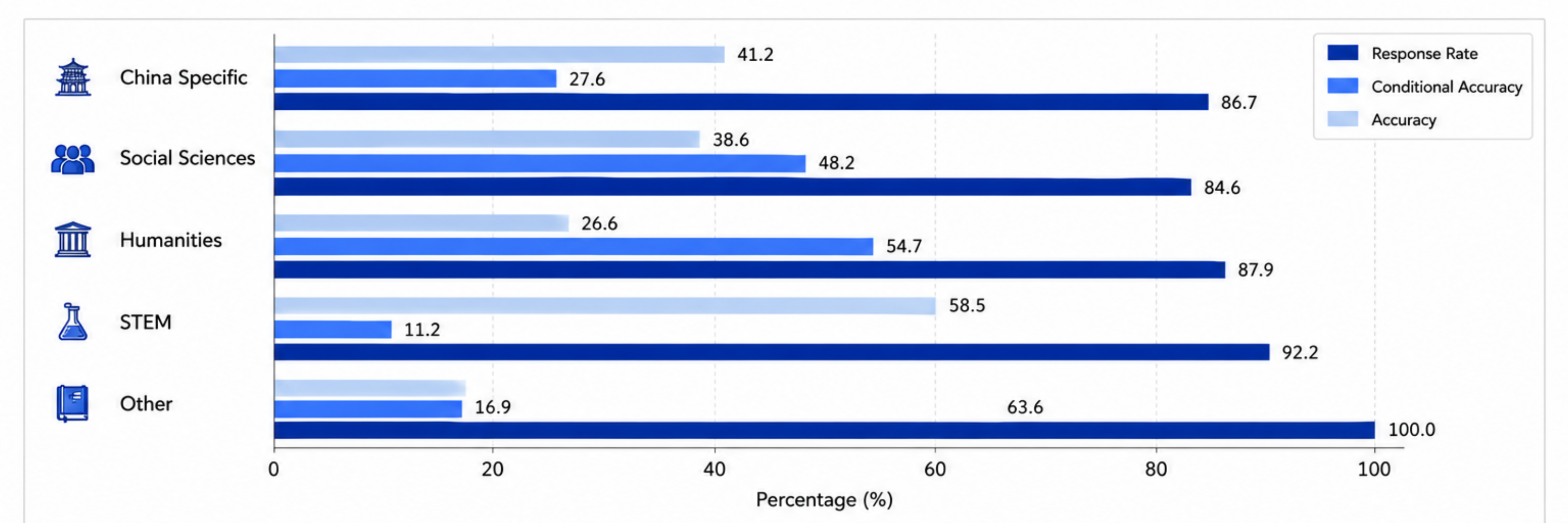}}
    \caption{The Highest Total Number of Indicators}
    \label{1st}
  \end{subfigure}
    \hfill
    \begin{subfigure}{0.8\linewidth}
    \centerline{\includegraphics[width=\columnwidth]{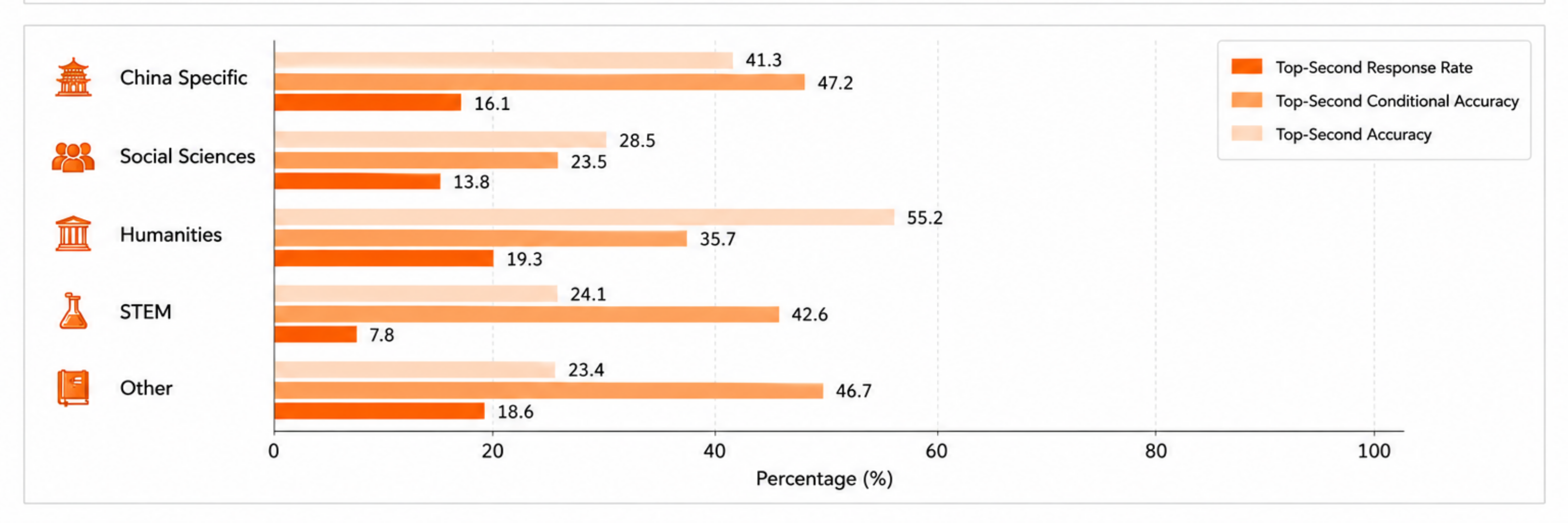}}
    \caption{The Second Total Number of Indicators}
    \label{2nd}
  \end{subfigure}
    \caption{Consistent Top-2 Indicator Rankings of \textit{Sun-Shine 1.0} across \textit{Ti-MMLU} Sub-benchmark}
    \label{1-f}
\end{figure}

As shown in Figure \ref{1-f}, \textit{Sun-Shine 1.0} ranks first and second based on the number of indicators, across all categories of the \textit{Ti-MMLU} sub-benchmark.

\clearpage

\subsection{Experiment on Tibetan Culture }
\label{appf2}

\subsubsection{Classical Chinese Generation}
\label{ccg}

\begin{figure}[ht]
\centering
\includegraphics[width=0.85\columnwidth]{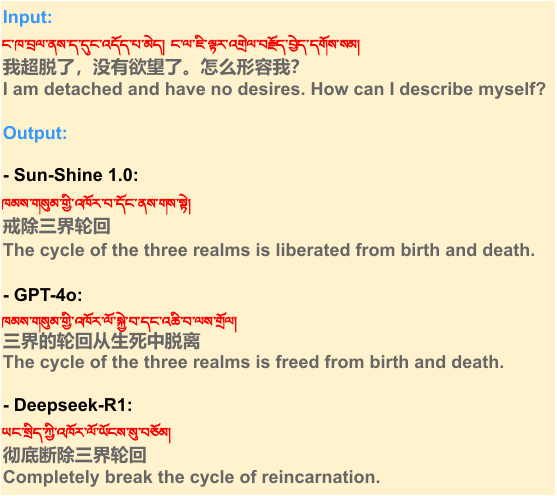}
\caption{Comparison of LLM Output} 
\label{dk-c}
\end{figure}

As shown in Figure~\ref{dk-c}, \textit{Sun-Shine 1.0} substantially outperforms DeepSeek-R1 \cite{r1} and GPT-4o \cite{gpt4o1} in classical Chinese. This advantage arises from training on \textit{TIBSTC}, which contains ancient poems, essays, and historical records. Manual revision preserves classical grammar, sentence structure, and style, resulting in outputs with a distinctly classical character, in contrast to the more modern style of DeepSeek-R1 \cite{r1} and GPT-4o \cite{gpt4o1}.

\subsubsection{Ancient Poetry Writing}
\label{apw1}

\begin{figure}[ht]
\centering
\includegraphics[width=0.9\columnwidth]{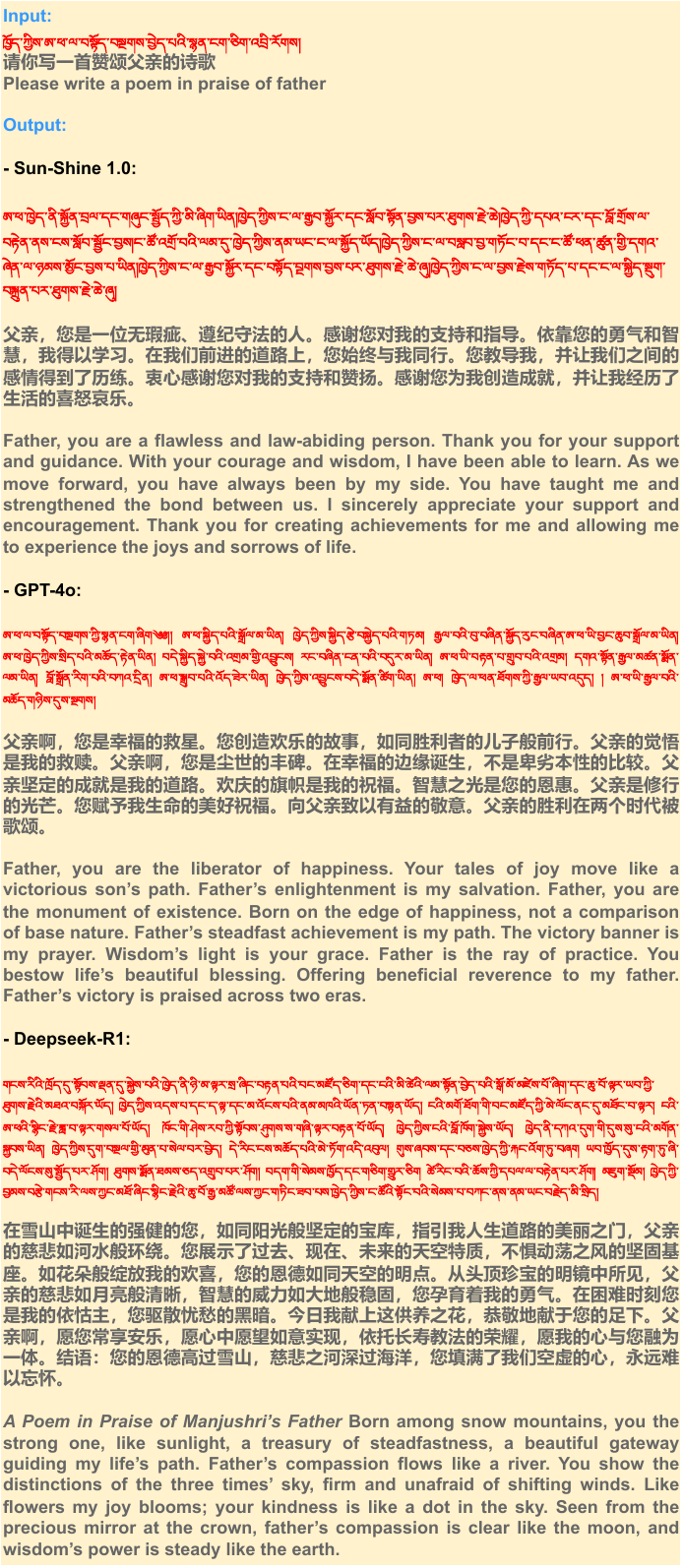}
\caption{Ancient Poetry Writing Generation} 
\label{apw}
\end{figure}

As shown in Figure~\ref{apw}, different LLMs exhibit clear stylistic differences when responding to the culturally sensitive prompt “Please write a poem in praise of father” across Tibetan, Chinese, and English. \textit{Sun-Shine 1.0} maintains a neutral and restrained tone, avoiding exaggerated praise or references to Tibetan classics. In contrast, GPT-4o \cite{gpt4o1} adopts an amplified, scripture-like style, while DeepSeek-R1 \cite{r1} frequently invokes the bodhisattva Mañjuśrī, resulting in more embellished and overfitted responses. Overall, \textit{Sun-Shine 1.0} outperforms DeepSeek-R1 \cite{r1} and GPT-4o \cite{gpt4o1} across most metrics, reflecting the impact of curated \textit{TIBSTC} data and human-in-the-loop training.

\subsection{Supplementary Experiment} 
\label{app-7}

We evaluate these LLMs on the Tibet Autonomous Region Tibetan–Chinese Translation 2024 test\footnote{\url{http://cn.zyw.xizang.gov.cn/}} (100-point TLP system). As shown in Table~\ref{test}, \textit{Sun-Shine 1.0} performs best on the more challenging A and B exams, which emphasize ancient Tibetan content, while achieving lower scores on the C and D exams that focus on modern Tibetan, including recent political topics. This pattern highlights \textit{Sun-Shine}’s strong proficiency in classical Tibetan and reflects the advantage of the \textit{TIBSTC} dataset, which contains a richer collection of ancient Tibetan texts.

\begin{table}[ht]
\centering
\scalebox{0.5}{
\begin{tabular}{l|c|c|c|c}
\hline
\textbf{LLM}  & \textbf{A} & \textbf{B}  & \textbf{C}  &  \textbf{D} \\
\hline
Sun-Shine  &  \textbf{63} & \textbf{61}  &44  &  42 \\
GPT-4o \cite{gpt4o1} & 45  & 47  & \underline{53}  & 51 \\
DeepSeek-R1 \cite{r1} & \underline{59}  & \underline{56}  &  \textbf{67}  & \textbf{66}\\
Qwen-2.5 \cite{qwen2.5} & 41  & 43  & \underline{53}  & \underline{54} \\
\hline
\end{tabular}}
\caption{Comparison of TLP Test}
\label{test}
\end{table}

\end{document}